\documentclass[journal]{IEEEtran}
\ifCLASSINFOpdf
\else
\fi

\usepackage{amsmath,amsfonts,amssymb}
\usepackage{graphicx}
\usepackage{multirow}
\usepackage[colorlinks=true, allcolors=blue]{hyperref}
\usepackage{bm}
\usepackage{hyperref}
\usepackage{cite}

\usepackage{tabularx}
\usepackage{booktabs}

\usepackage{marginnote}

\begin{document}
%
\title{Deep Learning based Switching Filter for\\ Impulsive  Noise Removal in Color Images}
%
%
%

\author{Krystian Radlak,
        Lukasz Malinski,
        and~Bogdan Smolka,~\IEEEmembership{Member,~IEEE}
\thanks{K. Radlak and Bogdan Smolka are with  Faculty of Automatic Control, Electronics and Computer Science, Silesian University 	of Technology, Akademicka 16, 44‐100 Gliwice, Poland, e-mail: krystian.radlak@polsl.pl, bogdan.smolka@polsl.pl}
\thanks{Lukasz Malinski is with  Faculty of Materials Engineering and Metallurgy, Silesian University
	of Technology, Krasinskiego 8,	40-019 Katowice, Poland, e-mail: lukasz.malinski@polsl.pl }
}

\maketitle

\begin{abstract}
Noise reduction is one the most important and still active research topic in low-level image processing due to its high impact on object detection and scene understanding for computer vision systems. Recently, we can observe a substantial increase of interest in the application of deep learning algorithms in many computer vision problems due to its impressive capability of automatic feature extraction and classification. These methods have been also successfully applied in image denoising, significantly improving the performance, but most of the proposed approaches were designed for Gaussian noise suppression.  In this paper, we present a switching filtering design intended for impulsive noise removal using deep learning. In the proposed method, the impulses are identified using a novel deep neural network architecture and noisy pixels are restored using the fast adaptive mean filter. The performed experiments show that the proposed approach is superior to the state-of-the-art filters designed for impulsive noise removal in digital color images.  
\end{abstract}


\ifCLASSOPTIONpeerreview
\begin{center} \bfseries EDICS Category: 5–ALGM \end{center}
\fi
%
\IEEEpeerreviewmaketitle

\section{INTRODUCTION}
\IEEEPARstart{I}{mage}

denoising is a long-standing research topic in low-level image processing that still receives much attention from computer vision community. Over the last three decades, 
a considerable increase in the effectiveness of algorithms took place
but despite these improvements, modern miniaturized high-resolution, low-cost image sensors still provide a limited quality, when operating in low lighting conditions. Therefore,  image enhancement and noise removal are very important operations of digital image processing.
\par In practice, we can observe various types of noise that significantly degrade the quality of captured images. One of them is the so-called 
\emph{impulsive noise}, which may appear due to electric signal instabilities, corruptions in physical memory storage, random or systematic errors in data transmission, electromagnetic interferences, malfunctioning or aging camera sensors, and poor lighting conditions \cite{Luk1,Bon1,LiuSzeliski2008,SmolkaMalikMalik2012,Malinski2019}.  This type of noise causes a total loss of information at certain image locations because the original color channels information is replaced by random values. 
\par In literature,  impulse noise is typically classified into  two main categories \cite{PlataniotisVenetsanopoulos2000,SmolkaPlataniotis2004,PhuTischer2007}.  The first one is the Channel Together Random Impulse (CTRI), in which a  pixel channel may be replaced with any value in the image intensity range. The second noise model is Salt \& Pepper Impulse Noise (SPIN), in which a corrupted pixel value is set to either the minimum or the maximum from a range of possible image values (so it is set to either $0$ or $255$ for an $8$-bit image).  In both models, the main parameter is the noise density $\rho$, which denotes the fraction  of corrupted pixels in the processed image. In this paper, we focus on the CTRI model, which  is more common and the proposed filter can be directly applied to the SPIN model.

\par    The classical method for removal of impulsive noise is the median filter. Generally, the median concept for color images is based on vector ordering, in which image pixels are treated as three-dimensional vectors. This gives better results than   processing  image channels independently \cite{LukacSmolka2005Entropy,LukacSmolka2005,LukacPlataniotis2006}. The basic example of filters utilizing the vector ordering concept is the  Vector Median Filter (VMF) \cite{AstolaHaavisto1990},
which effectively removes  the impulses, but fails when noise density is very high and impulses are grouped in clusters, which are retained forming colorful blotches. 

Another drawback of filters based on vector ordering is the fact that every pixel of the image is processed, regardless of whether it is contaminated or not. This may result in strong signal degradation and introduces a visible blurring effect, especially in highly textured regions.    In many applications, it may become an undesired property and therefore a plethora of improvements have been proposed in literature \cite{Luk9,Luk1,Luk2,Luk3,CelebiKingravi2007,SmolkaPerczak2007,SmolkaMalikMalik2012}.
To preserve image details and still efficiently suppress impulsive noise, a family of filters based on fuzzy set theory was introduced, in which a combination of impulsive noise detection and a replacement scheme based on averaging is performed \cite{Cha1,PlataniotisAndroutsos1999,ShenBarner2004,Mor5,Camarena-2008,MorillasGregori2009}. However, these methods still may alter clean pixels in the processed image.
\par
\begin{figure} [ht]
	\begin{center}
		
		\includegraphics[width=0.48\textwidth]{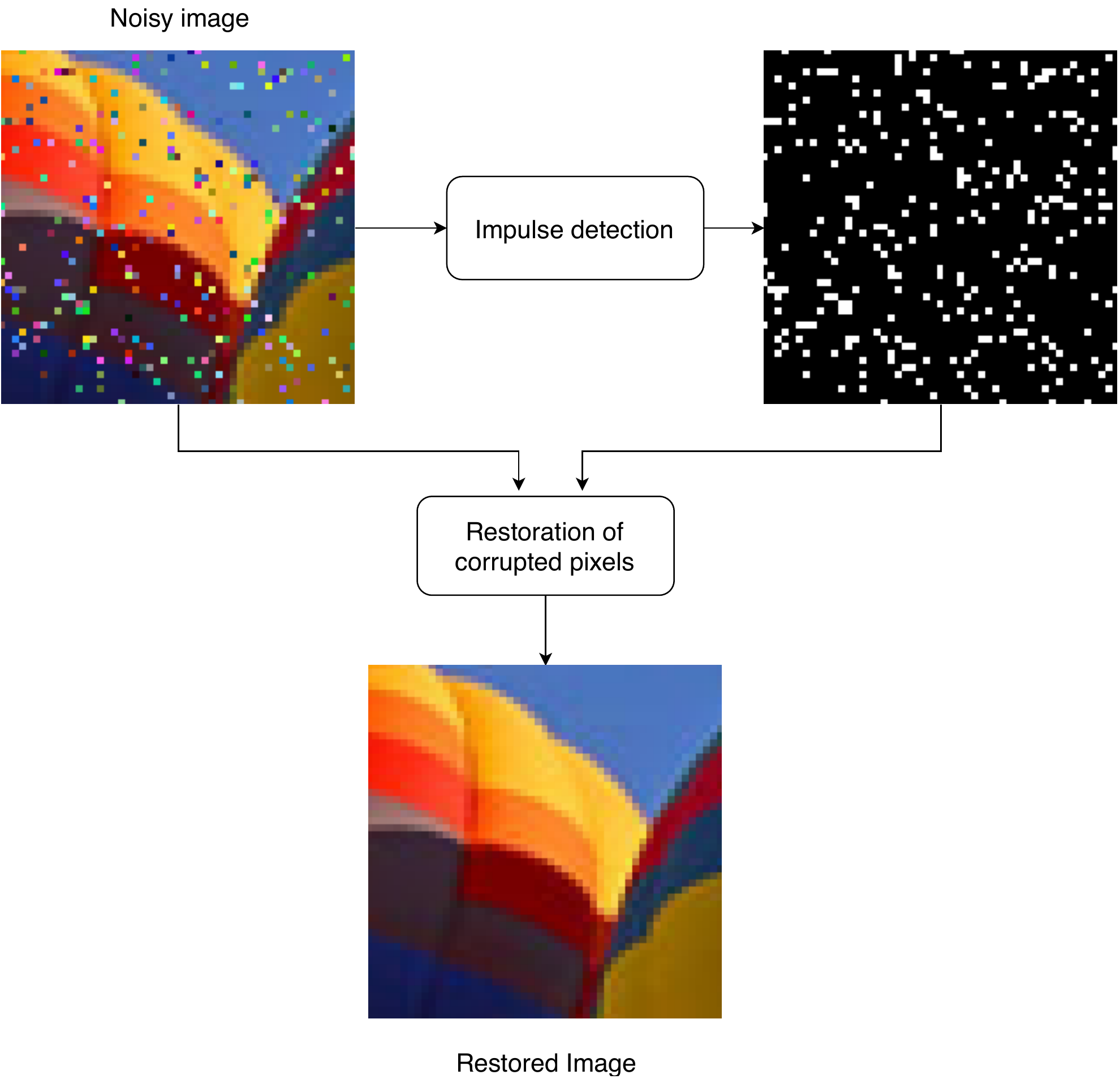}
		
	\end{center}
	\caption{ A general scheme of a switching filter. \label{fig:switching_filter} }
	
\end{figure}

 An effective approach to retain uncorrupted pixels is based on the switching concept\cite{Mal2}. A general scheme of switching filter is presented in Fig. \ref{fig:switching_filter}. In the majority of the switching techniques, it is necessary to determine the measure of dissimilarity between the processed color pixels and a threshold value that allows to classify the pixels as clean or distorted. One of the most popular measures of similarity used in switching filters is the ROAD (Rank-Ordered Absolute Differences) statistic introduced in \cite{GarnettHuegerich2005}, in which the trimmed cumulative distance of the pixels to their neighbors is utilized as a measure of pixel corruption.

\par Among switching filters, an important group of methods is based on the concept of a {\it peer group} \cite{DengKenney1999,KenneyDeng2001,Mal1}, in which the membership of a central pixel of the filtering window to its local neighborhood is determined in terms of the number of close pixels.

\par Another efficient family of switching filters 
utilizes the  elements of quaternion theory  \cite{Jin5,Geng-2012,Wang-2014}. In this concept, instead of the  commonly used Euclidean distance in a chosen color space, the similarity between pixels is defined in the quaternion form. 

\par Besides model-based methods, some switching filters also use classical machine learning approaches for impulses detection such as Support Vector Machines \cite{LinTzu2012} and fully connected neural networks  \cite{LiangLu2008,Kaliraj2010,Nair2013,Turkmen2016}. The detected impulses are restored using a median of uncorrupted pixels \cite{Nair2013,LiangLu2008}, adaptive and iterative mean filters \cite{Kaliraj2010} or the edge-preserving regularization method \cite{Turkmen2016}.

\par Recently, thanks to easy access to large image datasets and  advances in deep learning, the Convolutional Neural Networks (CNN) have led to a series of breakthroughs in various computer vision problems such as image segmentation, object recognition and detection. Concurrently, CNN have been also successfully applied for image denoising, focusing on the problem of Gaussian noise suppression. The recently proposed approaches significantly outperform classical model-based algorithms in terms of filtering efficiency and speed \cite{Lefkimmiatis2017,Lefkimmiatis2018,ZhangZuo2018}. However, despite the fact that 
image denoising using deep learning for Gaussian noise removal has been well-studied, little work has been done in the area of impulsive noise removal \cite{Fu2018,AmariaMiyazakia2018,RadlakMalinski2019}.

\par In one of the most promising approaches,  called  Denoising Convolutional Neural
Network (DnCNN),\cite{ZhangZuo2017}  the authors proved that residual learning and batch normalization are particularly beneficial in the case of Gaussian noise model.  Unfortunately, the network based on residual learning formulation is not effective in the case of  other types of operations like JPEG artifacts removal, deblurring or image resolution enhancement \cite{Zuo2018} and also in case of impulsive noise. Applying residual learning for images contaminated by impulsive noise causes  all pixels in the image to be altered,
even those that were impulse-free, introducing unpleasant visual artifacts. 
Therefore, in this paper, we propose a modified version of the DnCNN called Impulse Detection Convolutional Neural
Network (IDCNN).  
\par In the proposed approach, in comparison to the basic DnCNN, we added a sigmoid layer to distinguish noise-free pixels from impulses and we reformulated the residual learning to the classification problem. In this way, the deep neural network is used as the impulse detector. Afterwards, the corrupted pixels are restored using an adaptive mean filter due to its good balance between simplicity and restoration effectiveness.



%
%
%

In summary, we make the following contributions:
\begin{itemize}
	\item we propose a switching filter that uses deep learning for detection of corrupted pixels and adaptive mean filter for their restoration,
	\item we introduce a neural network architecture for impulse detection in images contaminated by impulsive noise,
	\item we analyze the impact of different network's parameters on impulse detection efficiency,
	\item we publish the source code of the proposed approach at \hyperlink{http://github.com/k-radlak/IDCNN}{http://github.com/k-radlak/IDCNN}.
\end{itemize}

The paper is structured as follows. Section II describes the proposed switching filter based on deep learning, focusing on the architecture of the proposed IDCNN and its ablation study. Next Section presents a comparison of the proposed technique with state-of-the-art filters designed for impulsive noise removal. Finally, discussion and conclusions are given in Section IV.

\section{IMPULSIVE NOISE DETECTION USING CONVOLUTIONAL NEURAL NETWORK}

Recently, application of deep learning for image denoising  received much attention from computer vision community  due to its significant performance improvement in comparison to the classical machine learning algorithms. One of the most interesting methods, inspired by VGG network \cite{SimonyanZ15a}, intended for Gaussian noise is the Denoising Convolutional Neural Network (DnCNN) introduced by Zhang et al. \cite{ZhangZuo2017}. 
\par The DnCNN contains a sequence of convolutional layers followed by Rectified Linear Unit (ReLU) \cite{KrizhevskySutskever2012} and Batch Normalization (BN) \cite{IoffeSzegedy2015}. The first layer has a convolution filter and ReLU activation. The second and each
consecutive layer consists of a convolution filter, BN and ReLU activation, except the last layer 
which uses only convolution. The training of the network is based on the concept of deep residual learning \cite{HeZhang2016}, in which the network does not estimate the original values of the undistorted image, but instead learns to estimate the difference between noisy and clean image. 
\par More formally, let us assume that  $Y$ denotes a noisy image that is an input of DnCNN, $X$ stands for a clean image and $\mathcal{R}(Y)$ is the output of the network, where  $Y=X + \mathcal{R}(Y)$. The estimation of $X$ can be formulated as $\hat{X} = Y - \mathcal{R}(Y)$. 

\par The DnCNN filter outperforms most of the state-of-the-art algorithms designed for Gaussian noise removal, but due to the fact that it uses  residual learning, the original DnCNN trained on impulsive noise model also alter  non-corrupted pixels as was shown in \cite{RadlakMalinski2019}. Therefore, in the proposed approach, we modified the original DnCNN architecture to ensure that noise-free pixels will be not affected and we introduced  IDCNN. 
\par
\begin{figure*} [ht]
	\begin{center}
		\begin{tabular}{c} 
			\includegraphics[width=0.92\textwidth]{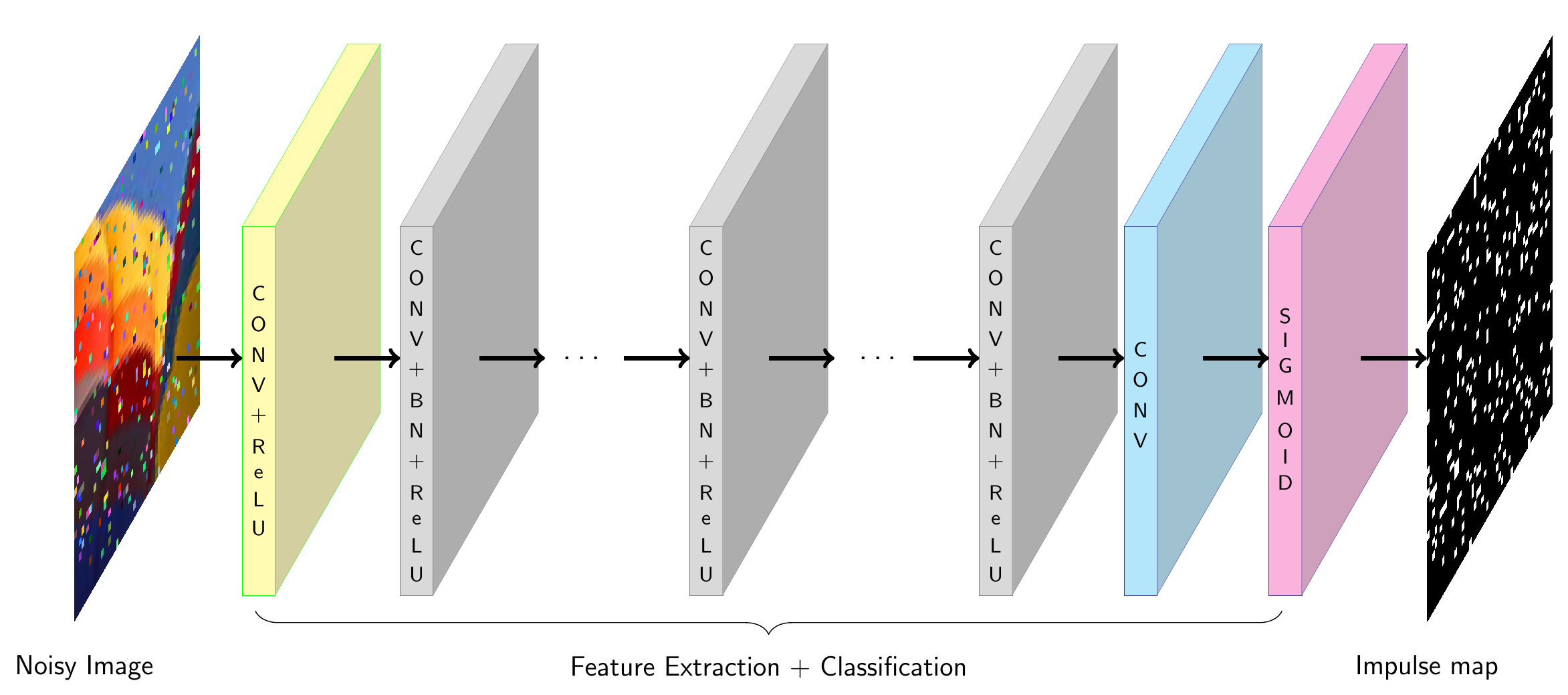}
		\end{tabular}
	\end{center}
	\caption{ The architecture of the proposed network for impulsive noise detection. \label{fig:dncnn} }
\end{figure*}

  In the proposed IDCNN, instead of the usage of residual learning, we employed all layers proposed in DnCNN for feature extraction and we added a sigmoid layer which estimates the probability of the pixel being an impulse or noise-free.  The architecture of the proposed network is depicted in Fig.~\ref{fig:dncnn}.
\par
\begin{figure*} [ht]
	\begin{center}
		\begin{tabular}{c} 
			\includegraphics[width=0.72\textwidth]{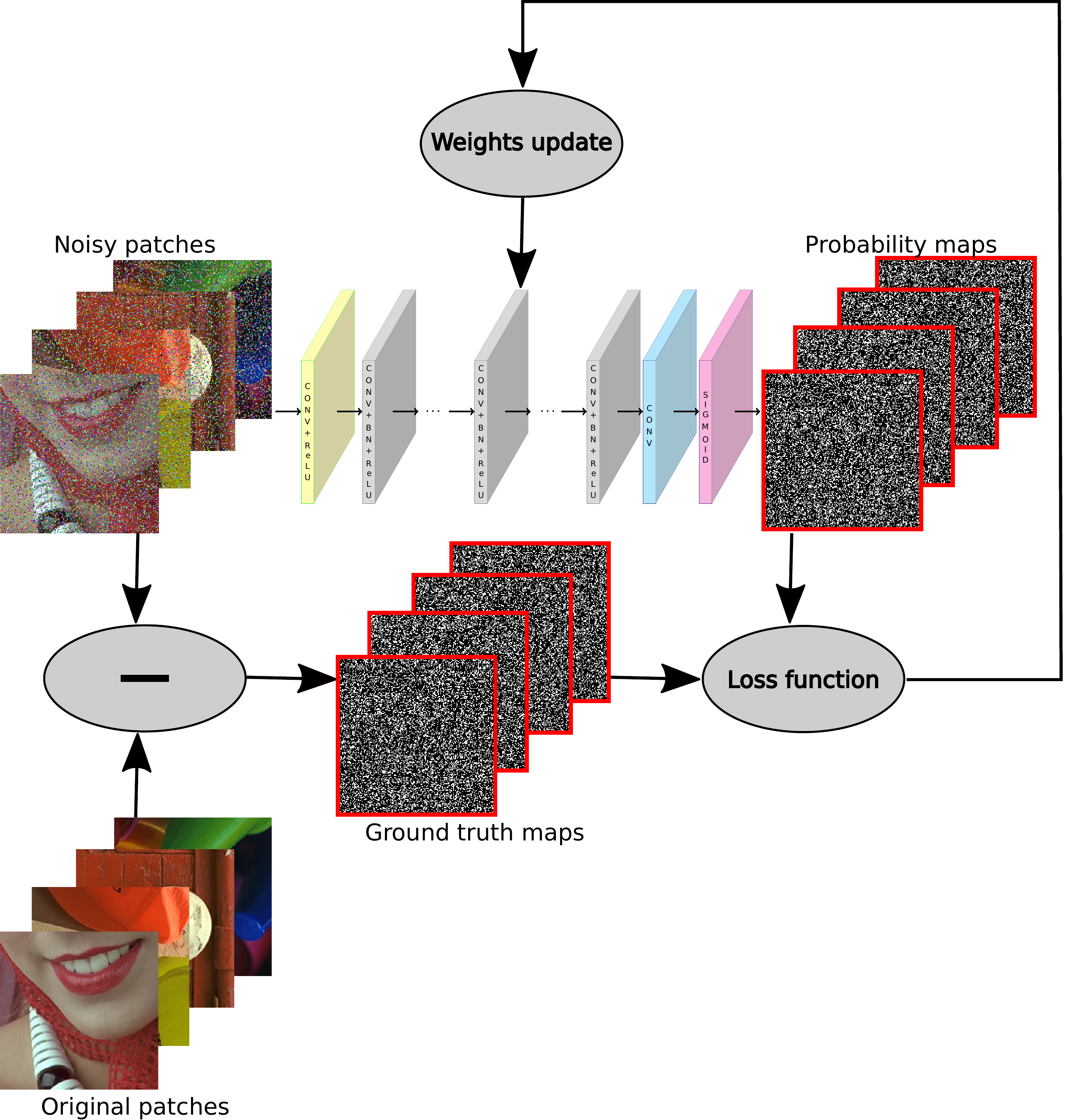}
		\end{tabular}
	\end{center}
	\caption[example] 
	{ \label{fig:dncnn_train} 
		Training of the proposed IDCNN detector.}
\end{figure*}

  The introduced architecture also required a change in the training procedure. In the original DnCNN, during training the images are divided into a small square and non-overlapping patches of size $p \times p$. The loss function is then calculated between clean patches and the denoised one. In our approach, we generate a ground truth noise map $M$ and in the training procedure we calculate the loss function between patches cropped from $M$ and the noise map $\hat{M}$, estimated by the network as shown in Fig. \ref{fig:dncnn_train}.  More formally, the loss function  is defined as follows:
\begin{equation}
L(\Theta)=\frac{1}{N}\sum_{i=1}^N (M_i -\hat{M_i})^2,
\end{equation}
where $\Theta$ denotes a set of trainable parameters, $M_i,\hat{M_i}$ 
denote for each patch $i$ the original and the estimated noise map respectively, and $N$ denotes the number of patches (small cropped images) used in the training.  Finally, the output of the IDCNN is a probability map that has to be binarized using a threshold to finally classify a pixel as either noisy or undistorted.

\par In order to restore the detected noisy pixels, we used the modified version of the adaptive arithmetic mean filter introduced in \cite{Kaliraj2010}, in which the authors proved that this approach gives satisfying  image quality and reasonable computational speed. This algorithm of restoration of the detected noisy pixel can be summarized as follows:
\begin{enumerate}
	\item Select initial window of size $W =  3 \times 3$ centered at the detected noisy pixel that should be replaced and calculate the number of pixels that are not corrupted by noise (using using an appropriate detection detector). If the number of uncorrupted pixels is lower than 1 then go to step 2 else go to step 4.
	\item  Increase the size of $W$ by 2.
	\item Calculate the number of uncorrupted pixels inside $W$ and if the number of uncorrupted pixels is lower than 1 then go to step 2 else go to step 4.
	\item  Replace the current pixel by the average of all uncorrupted pixels inside $W$.
\end{enumerate}

The proposed algorithm should be applied to restore all pixels that were classified by the network as impulses. However, it is worth mentioning
here that the corrupted pixels can be replaced using other more robust techniques, e.g. image restoration algorithm based on deep neural network introduced in \cite{ZhangZuo2017IRCNN}. This issue will be the subject of follow-up research.


%
%

\subsection{Ablation study}
In order to evaluate the performance of the proposed network, we performed several experiments. We started with the default parameters that were proposed for DnCNN \cite{ZhangZuo2017}. These parameters are summarized in Tab. \ref{tab:parameters}.

\begin{table}[]\scriptsize
	\caption{Summary of the network parameters.}
	\label{tab:parameters}
	\centering
	\begin{tabular}{ll} \hline
		Parameter & Value  \\ \hline
		Number of convolutional layers & 17 \\
		Number of filters in convolutional layer  &  64 \\
		Size of convolutional window &  $3\! \times \! 3$ \\
		Number of epochs & 50\\
		Learning rate & 0.001 \\
		Learning rate decay & 0.1\\
		Epoch in which learning rate decay is used & 30\\
		Batch size & 128 \\
		Weights initialization  & Glorot uniform initializer\cite{Glorot2010} \\
		Weights optimization & ADAM optimizer\cite{KingmaBa2015}\\   
		Patch size in the training & $41\! \times \! 41$\\ \hline
	\end{tabular}
	
\end{table}

Additionally, for training purposes, all images were resized using bicubic interpolation in four scales 
$\{1, 0.9, 0.8, 0.7 \}$ 
and we  applied simple data augmentation: image 
rotations (90$^{\circ}$, 180$^{\circ}$, 270$^{\circ}$) and image flipping in the up-down direction. Here it is worth mentioning that small patches were used only in the training phase, but in inference,  the obtained convolution masks were applied to the whole image at once.

\par
\begin{figure*} [ht]
	\begin{center}
		\begin{tabular}{c} 
			\includegraphics[width=0.92\textwidth]{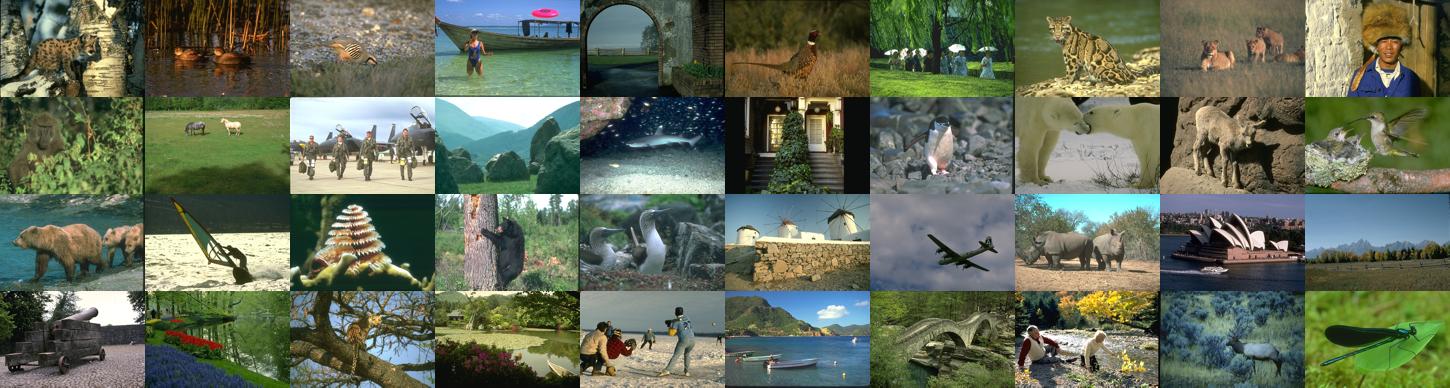}
		\end{tabular}
	\end{center}
	\caption
	{ \label{fig:bsd500} 
		Example images from Berkeley segmentation dataset (BSD500) \cite{ArbelaezMaire2011}.}
	
\end{figure*}

\begin{figure*} [ht]
	\begin{center}
		\begin{tabular}{c} 
			\includegraphics[width=0.92\textwidth]{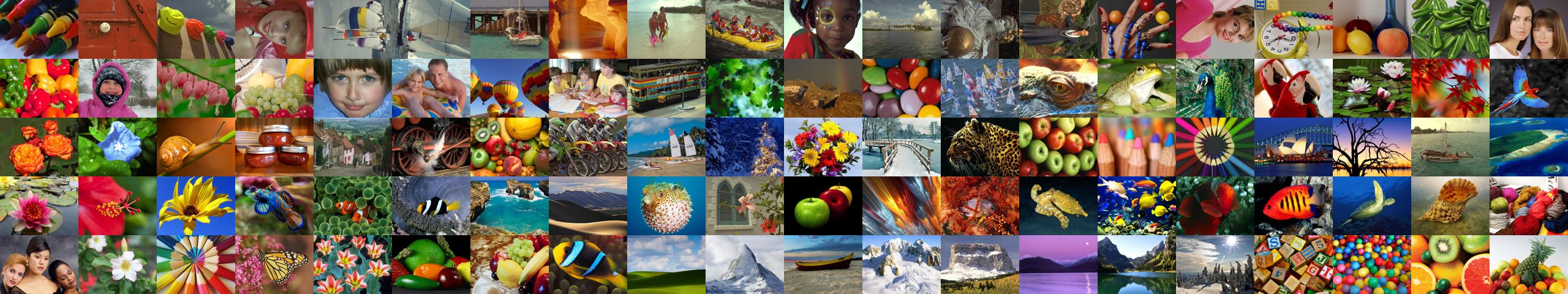}
		\end{tabular}
	\end{center}
	\caption
	{ \label{fig:benchmark} 
		Test dataset made available by Malinski and Smolka \cite{Mal1}.}
\end{figure*}

 In our experiments, similarly to the authors of the DnCNN approach, we used Berkeley segmentation dataset (BSD500) \cite{ArbelaezMaire2011} that consists of 500 natural images in resolution $481 \! \times \! 321$. 
Example images from BSD500 are depicted in Fig. \ref{fig:bsd500}.
For testing purposes, we used dataset introduced in \cite{Mal1} consisting of $100$ color images in resolution $640 \! \times \! 480$, which are presented in Fig. \ref{fig:benchmark}. In this paper, all presented results were obtained on this dataset.

For training purposes, these images were contaminated using the CTRI model. In this model each pixel $\mathbf{x_i} \in X, i=1,2, \ldots, Q$ is contaminated with probability $\rho$ and each channel obtains a new value $v_q, q=1,2,3$ from the range $[0,255]$ and drawn from a uniform distribution. This model can be formally defined as
\begin{equation}
\mathbf{y_i}  = \left\{
\begin{array}{ll}
(v_1,v_2,v_3)& \textrm{with probability }\rho,\\
\mathbf{x_i} & \textrm{with probability } 1-\rho,
\end{array}
\right.
\end{equation}
where $\mathbf{y_i}$ denotes a noisy image pixel.

In order to evaluate the noise detection efficiency of the proposed network and impact of the network's parameters, we propose to transform the problem into classification domain, instead of using traditional measures for image denoising. In the proposed evaluation methodology,  
the result of noise detection is represented by 
the estimated noise map  $\hat{M}$ and is compared to the ground truth map $M$. Then, impulsive noise detection problem can be transformed into noisy vs. clean pixels classification and the results can be presented using a confusion matrix that reports the number of True Positives (TP), True Negatives (TN), False Positives (FP), False Negatives (FN). For the impulse detection problem
\begin{itemize}
	\item  TP 
	are pixels that were correctly recognized as impulses,
	\item TN 
	are pixels that were correctly recognized as not being contaminated,
	\item FP 
	are pixels  that were  incorrectly  classified  as  noisy,
	\item FN 
	are pixels that were incorrectly classified as uncorrupted.
\end{itemize}

Finally, the network performance can be evaluated using weighted accuracy (wACC)  defined as 

$$
\textrm{wACC}= \rho\frac{\textrm{TP}}{\textrm{TP}+\textrm{FN}} + (1-\rho)\frac{ \textrm{TN}}{ \textrm{TN} + \textrm{FP} }.
$$
Weighted accuracy summarizes how many pixels were correctly classified when the classes are unbalanced and their cardinalities depends on selected noise intensity $\gamma$, but this metric does not distinguish what type of errors we made if we miss an impulse or incorrectly classify a clean pixel as noisy. 
\par In statistical hypothesis testing, two types of errors are defined.  Type I error occurs when a true null hypothesis is incorrectly rejected and it is also known as False Positive. A type II error occurs when the null hypothesis is false, but erroneously fails to be rejected and it is known as False Negative. Therefore in this work, we used two additional metrics that allow evaluating the portion of both types of errors that are made by the proposed detector. First of them is the False Positive Rate (FPR) defined as
$$
\textrm{FPR}= \frac{\textrm{FP}}{\textrm{FP} + \textrm{TN}},
$$
which shows the ratio of incorrectly classified clean pixels as impulses to the total number of clean pixels in the processed image. The second of them is the False Negative Rate (FNR) defined as
$$
\textrm{FNR}= \frac{\textrm{FN}}{\textrm{TP} + \textrm{FN}},
$$
which shows the 
ratio of incorrectly detected noisy pixels to the total number of impulses in an image. To evaluate the network performance on the whole test set, we determined wACC, TPR and FPR for each image and the average for all images from the test set was calculated.
\par
\begin{figure} [ht]
	\begin{center}
		\begin{tabular}{ccc} 
			\includegraphics[width=0.22\textwidth]{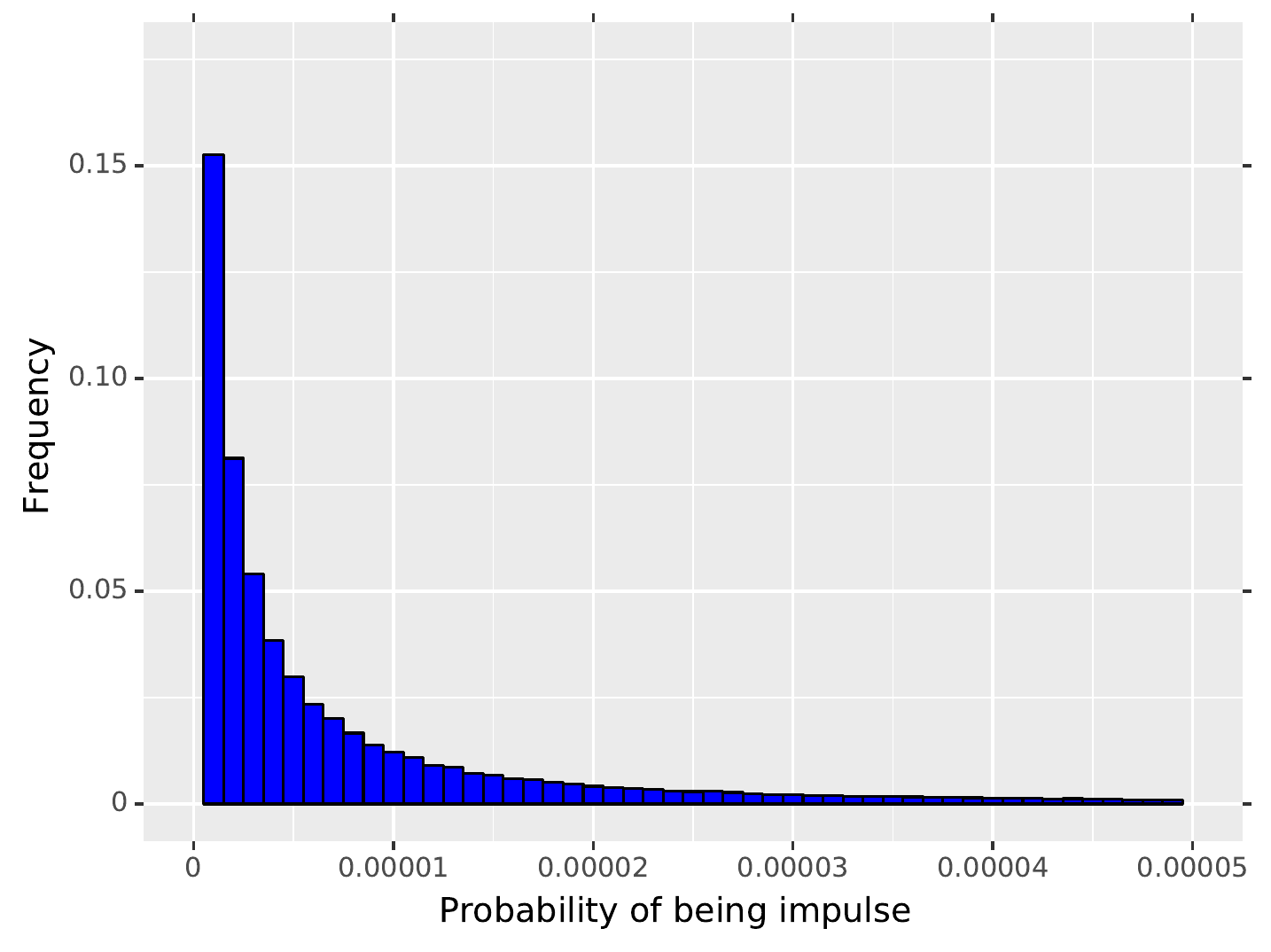} &
			\includegraphics[width=0.22\textwidth]{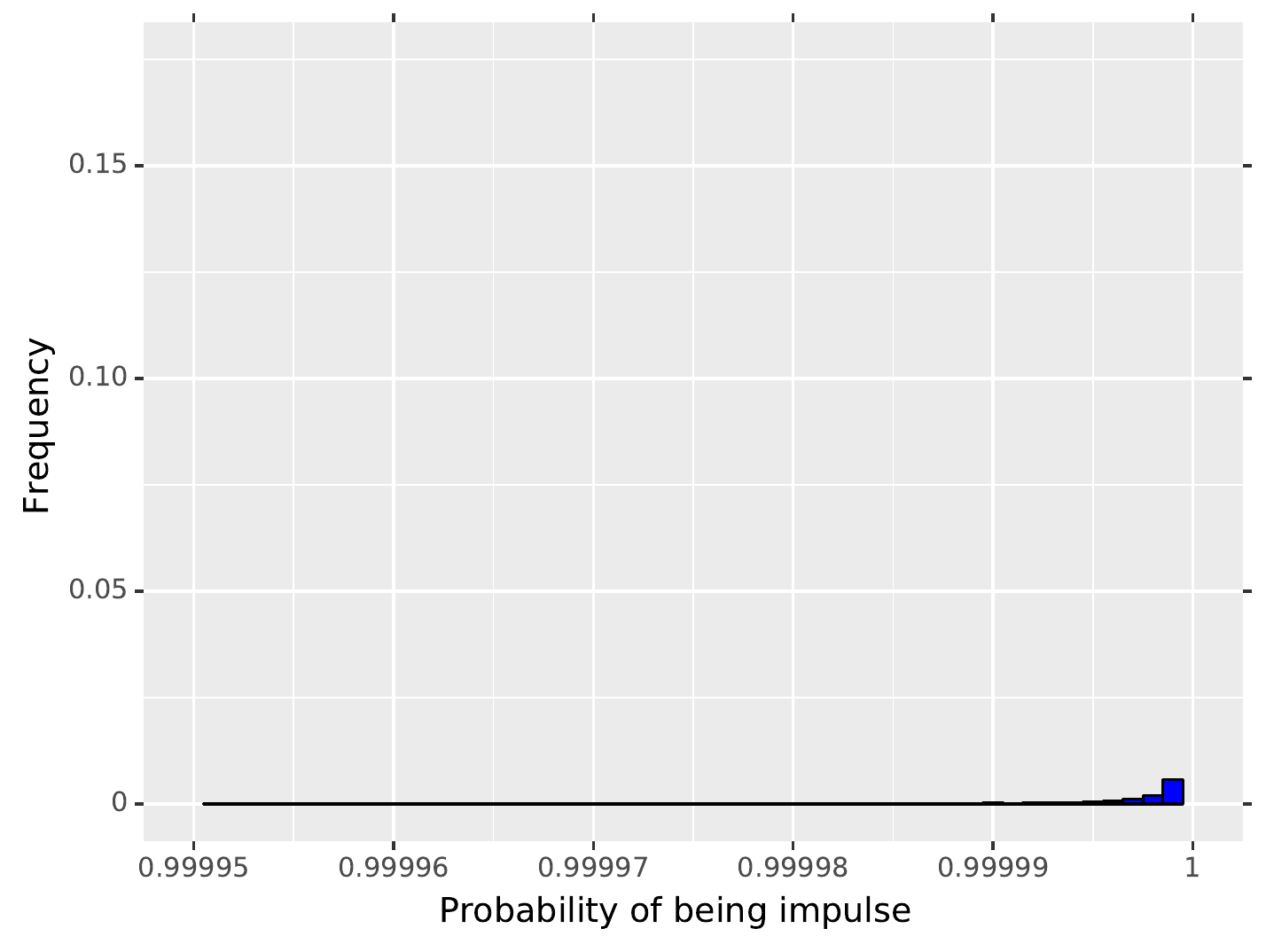} \\
			\multicolumn{2}{c}{Image contaminated with $\rho=0.1$ } \\
			
			\includegraphics[width=0.22\textwidth]{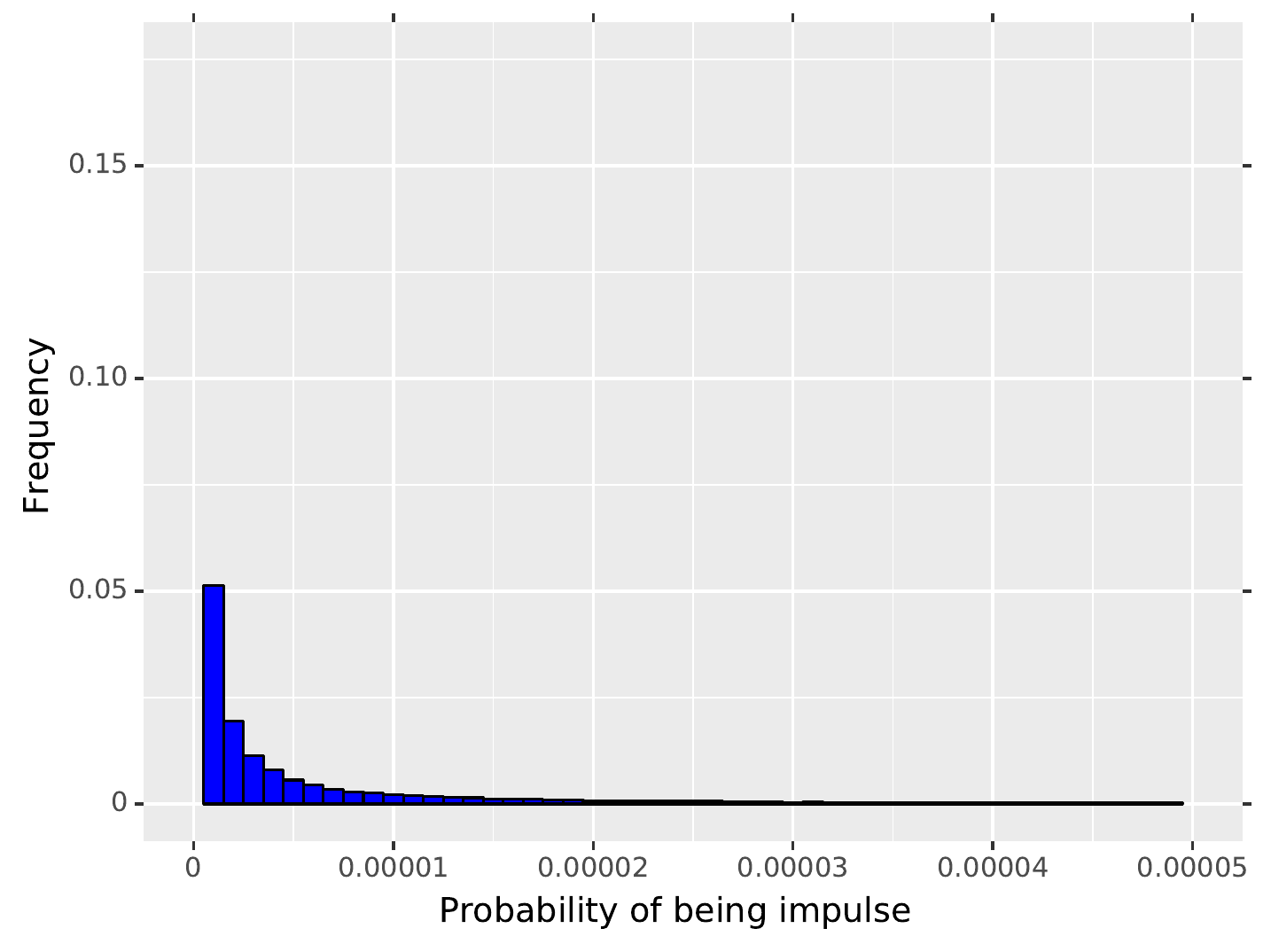} &
			\includegraphics[width=0.22\textwidth]{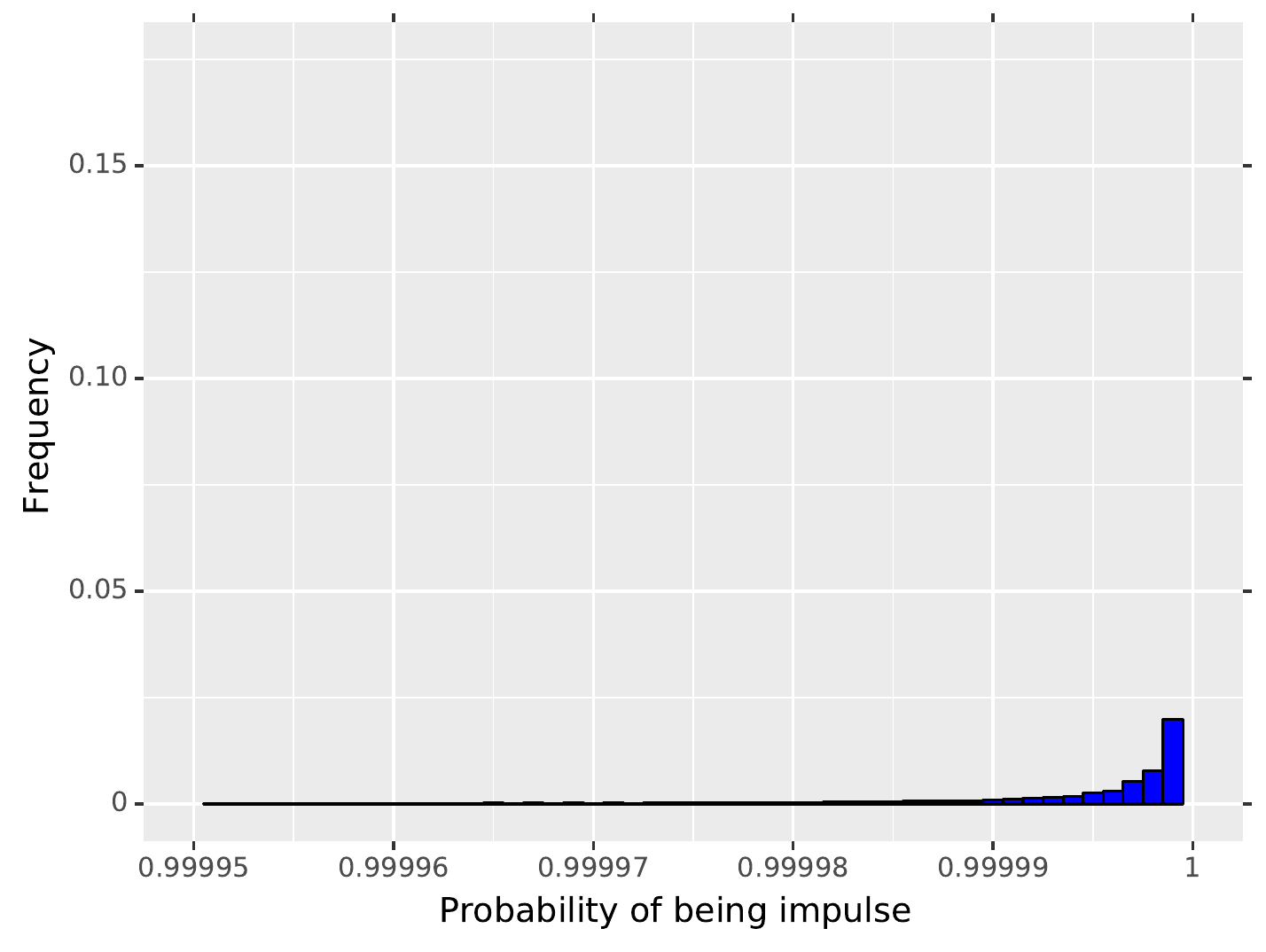} \\
			\multicolumn{2}{c}{Image contaminated with $\rho=0.5$ }
		\end{tabular}
	\end{center}
	\caption[example] 
	{ \label{fig:histogram} 
		Noise probability map distributions estimated by the proposed IDCNN for the example
		image.}
\end{figure}

  To correctly localize impulses in the image based on the output of the proposed IDCNN, in the first step it is necessary  to estimate the proper value of the threshold to select, which pixels are contaminated by impulsive noise. Selection of the optimal threshold typically can significantly affect the final results, but we noticed that the values of the probabilities returned by the network are close to 0 if a pixel is clean and close to 1 if a pixel is classified as an impulse. Example histograms with 
distributions of probabilities that a pixel is an impulse returned by IDCNN are presented in Fig. \ref{fig:histogram} 
(on the left we present only a selected part of the histogram and the plot, on the right we show parts of the same histogram but with a different range on the x-axis).
Therefore, in our research, we set the probability threshold to 0.5 as it does not have any impact on the classification results. 

\par
\begin{figure*} [ht]
	\begin{center}
		\begin{tabular}{ccc} 
			\includegraphics[width=0.3\textwidth]{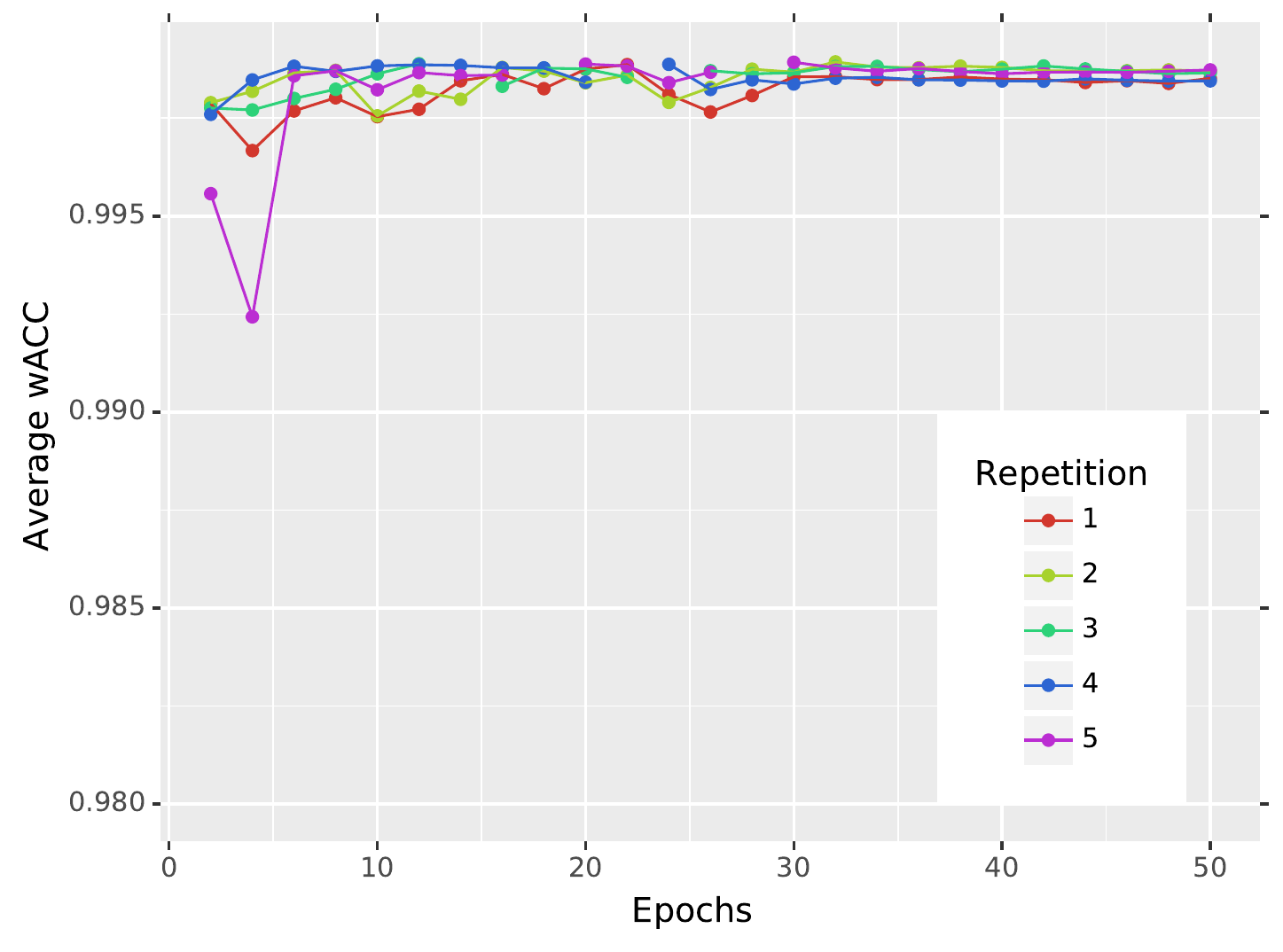} &
			\includegraphics[width=0.3\textwidth]{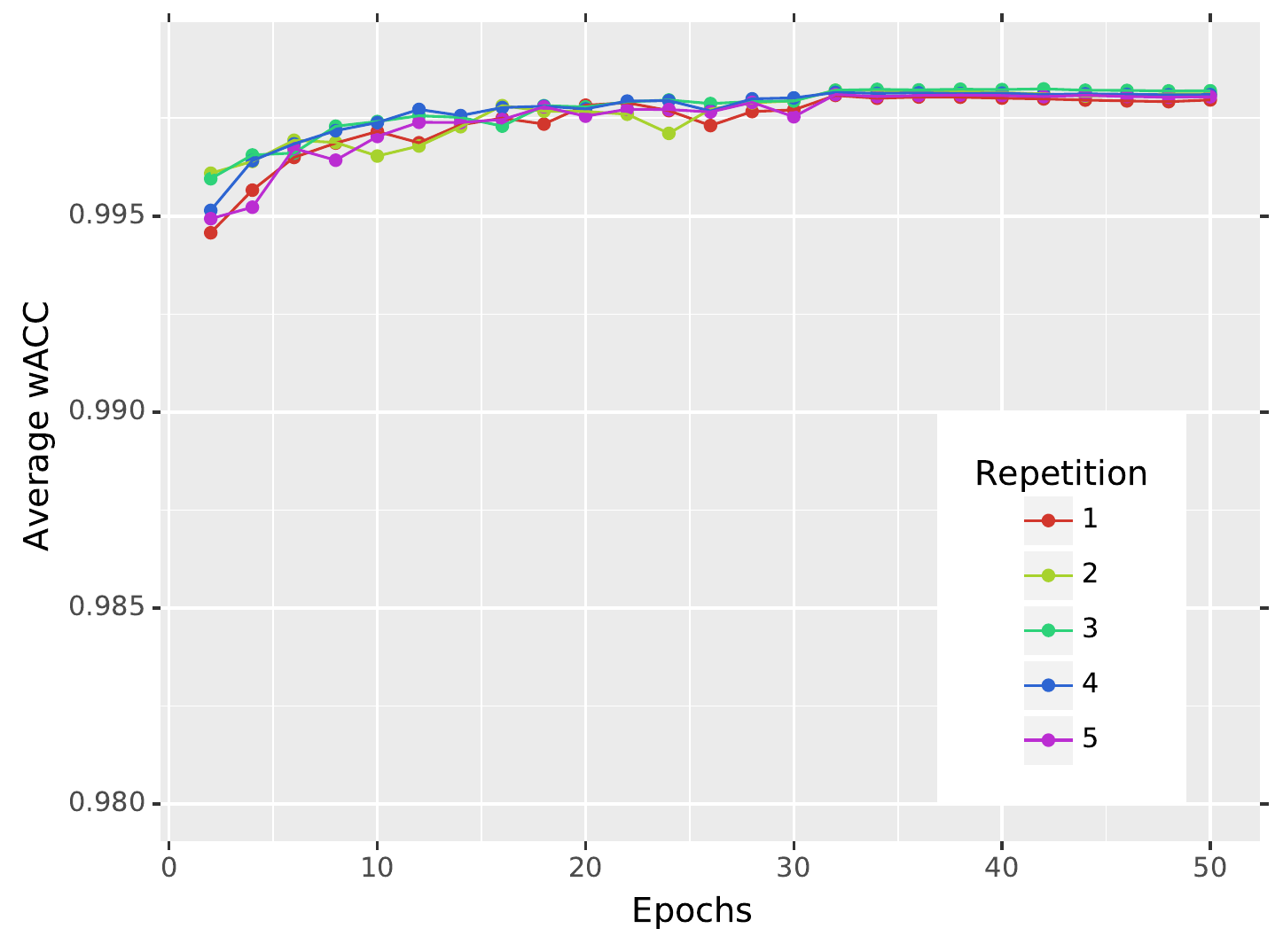} &
			\includegraphics[width=0.3\textwidth]{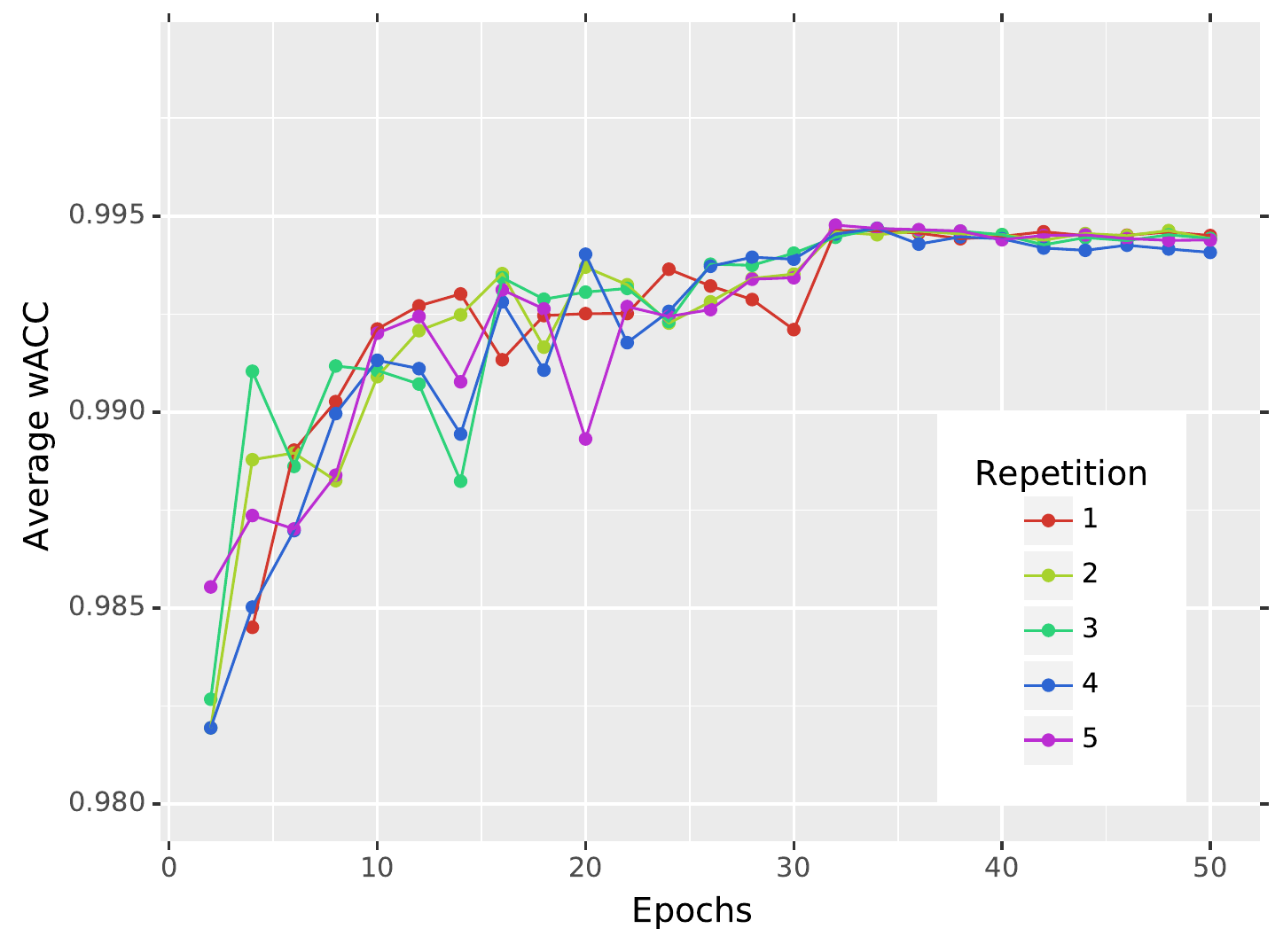} \\

			\includegraphics[width=0.3\textwidth]{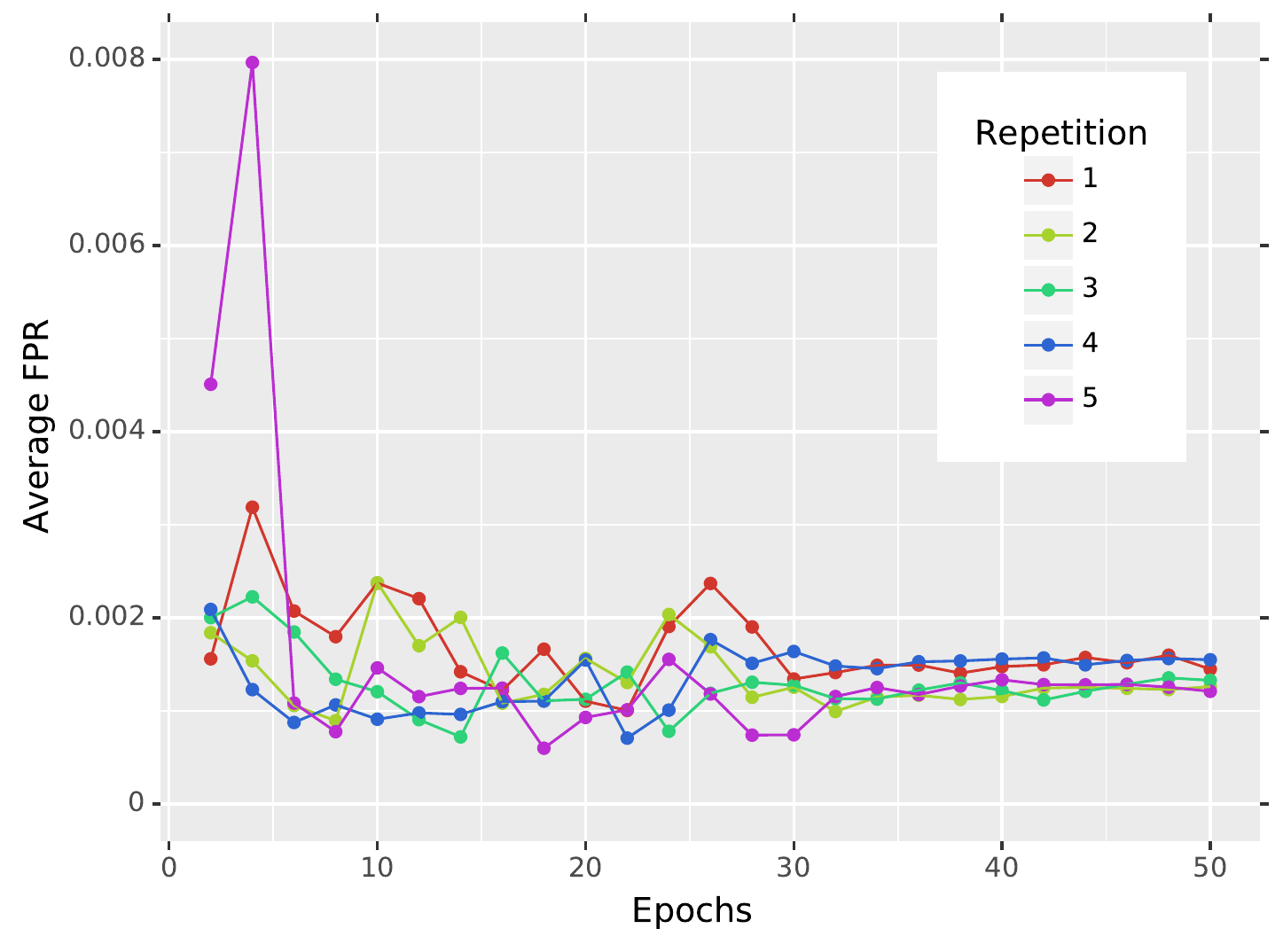} &
			\includegraphics[width=0.3\textwidth]{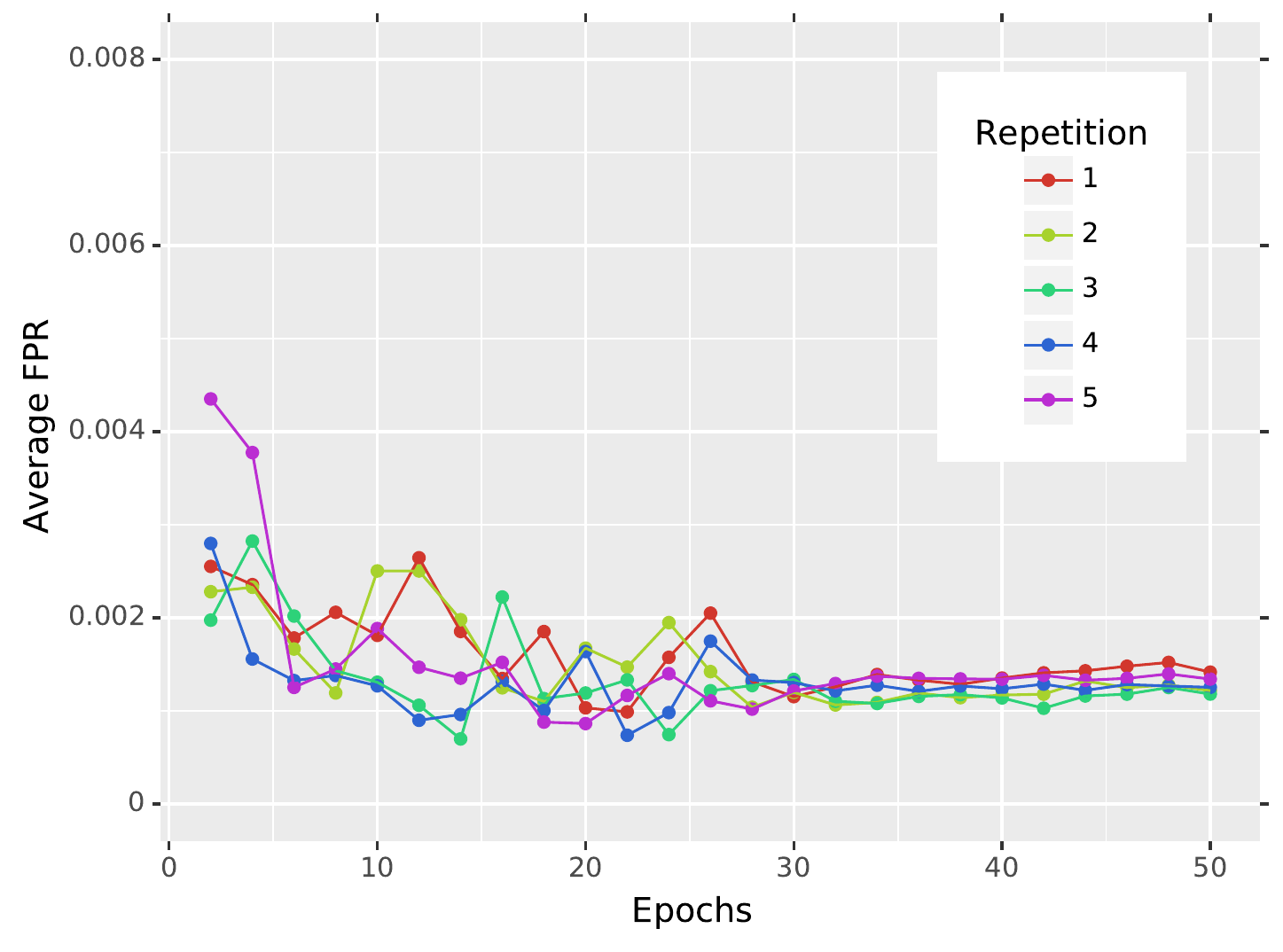} &
			\includegraphics[width=0.3\textwidth]{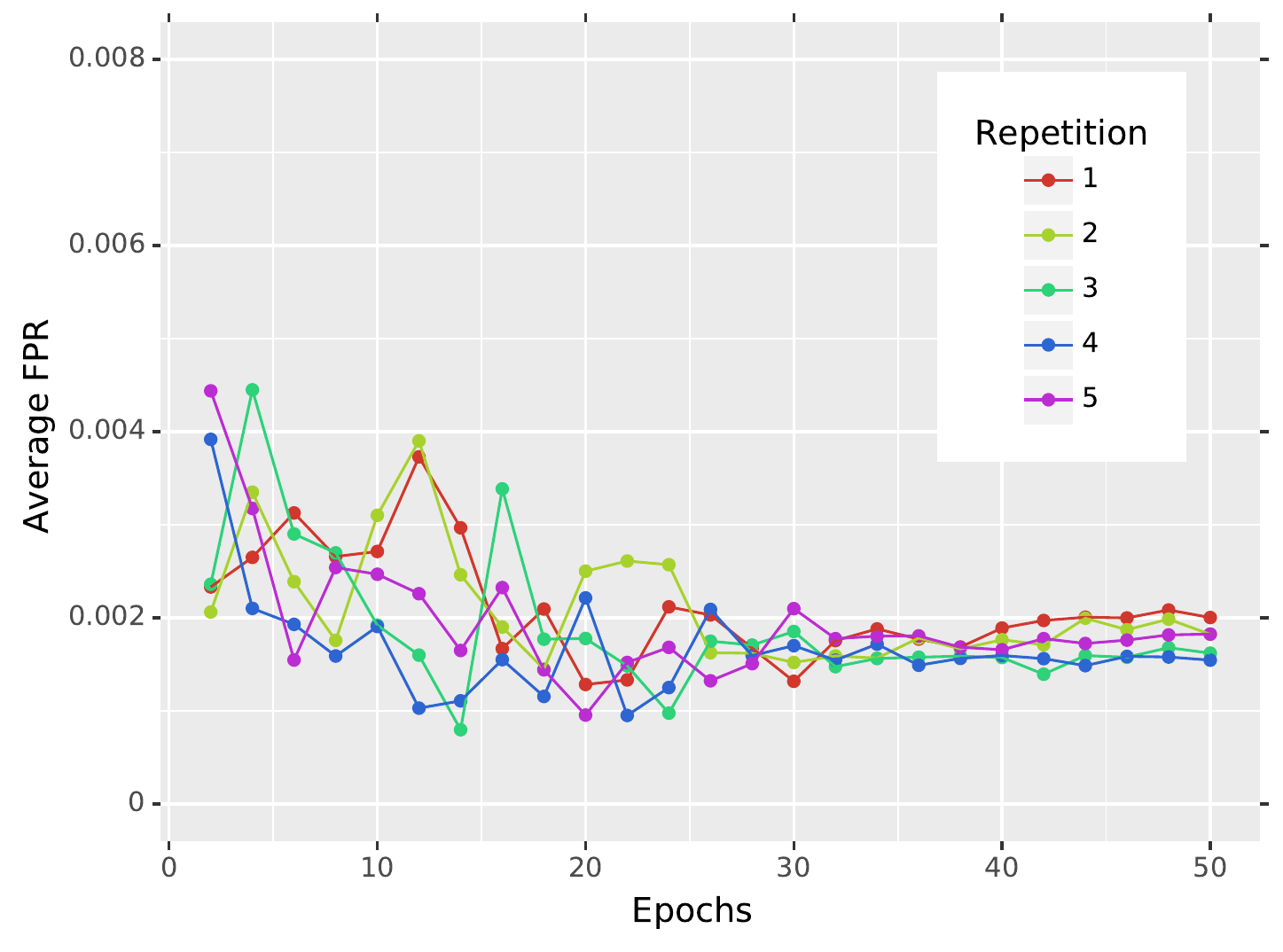} \\
			
			\includegraphics[width=0.3\textwidth]{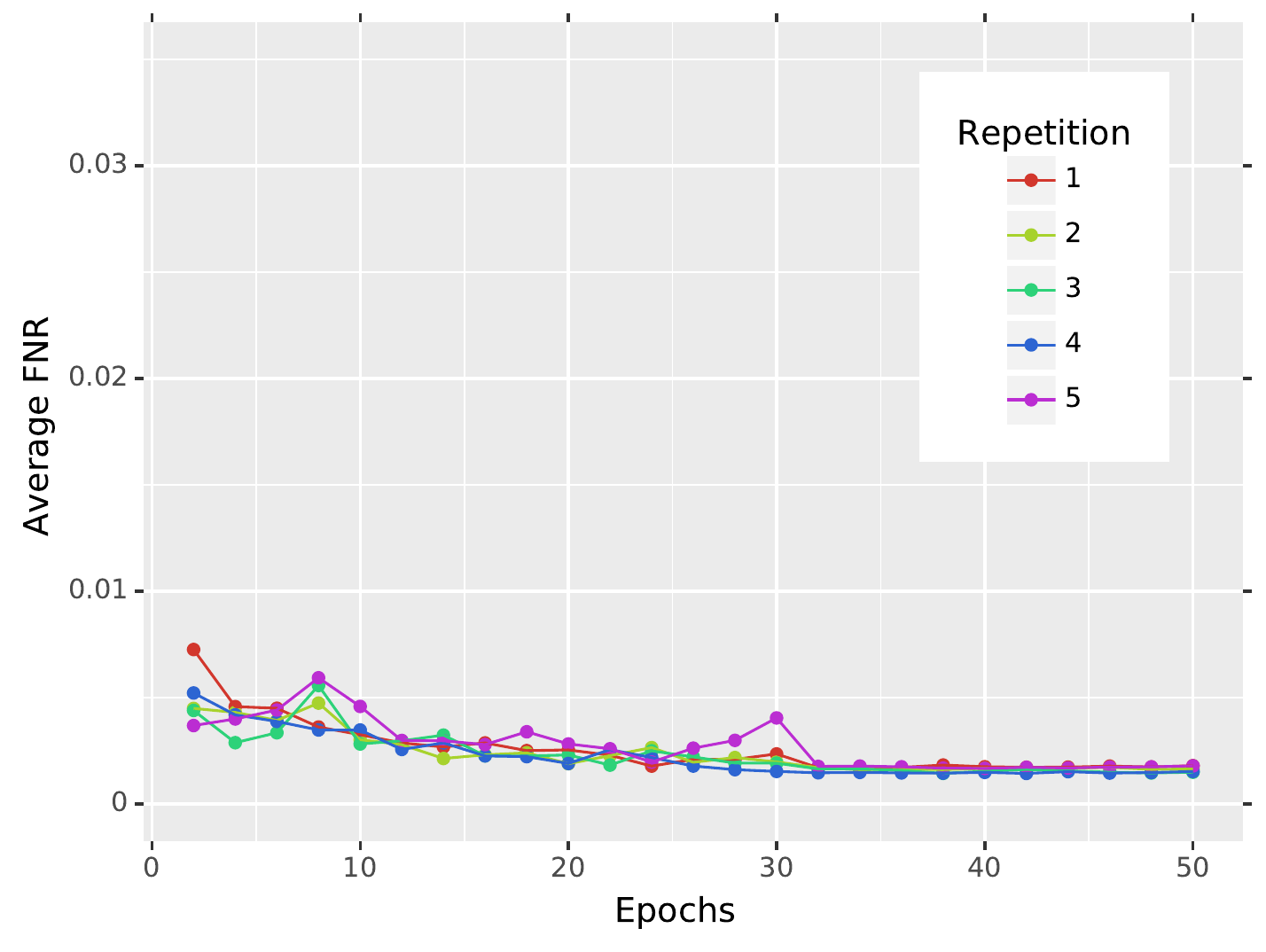} &
			\includegraphics[width=0.3\textwidth]{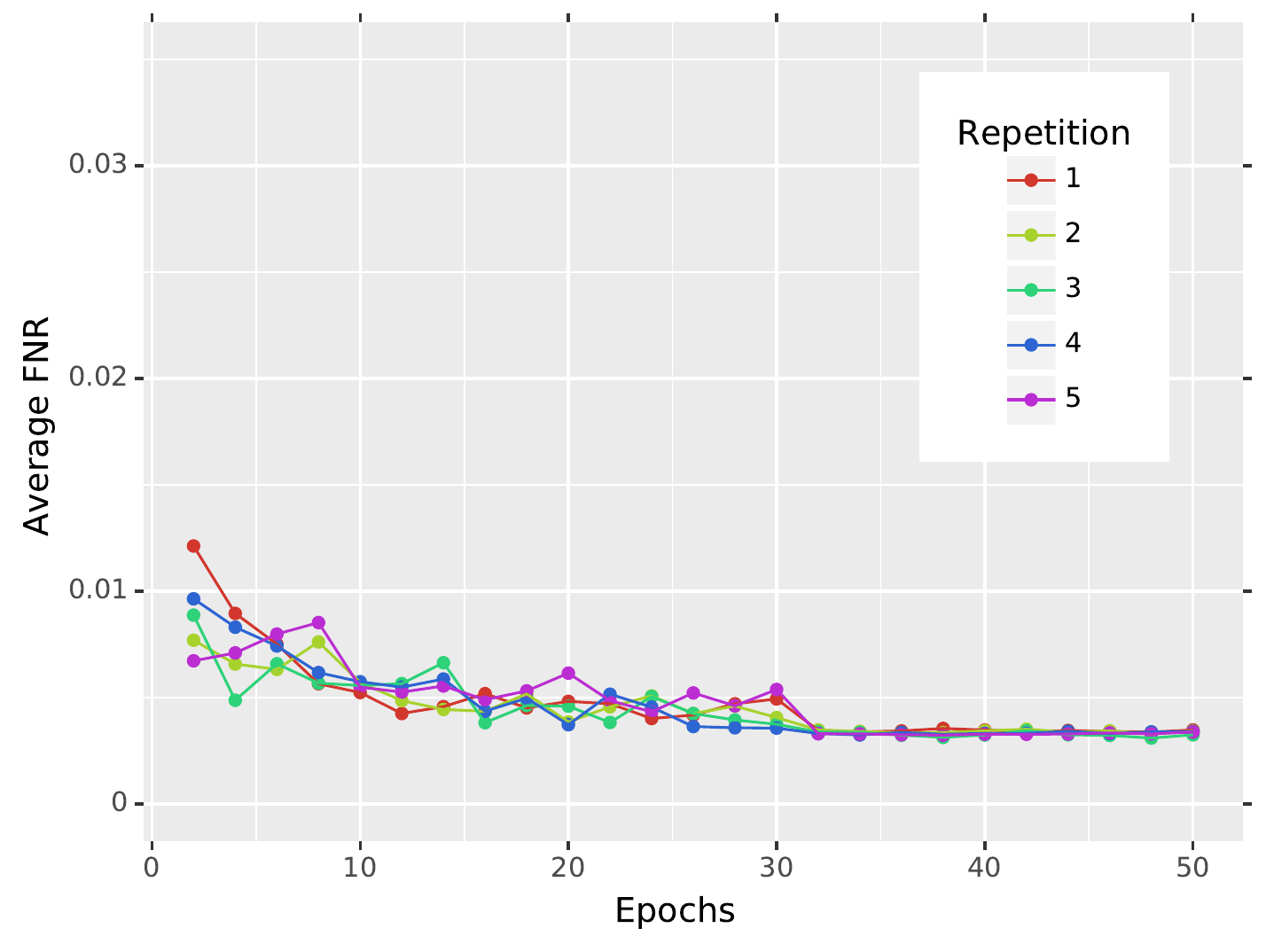} &
			\includegraphics[width=0.3\textwidth]{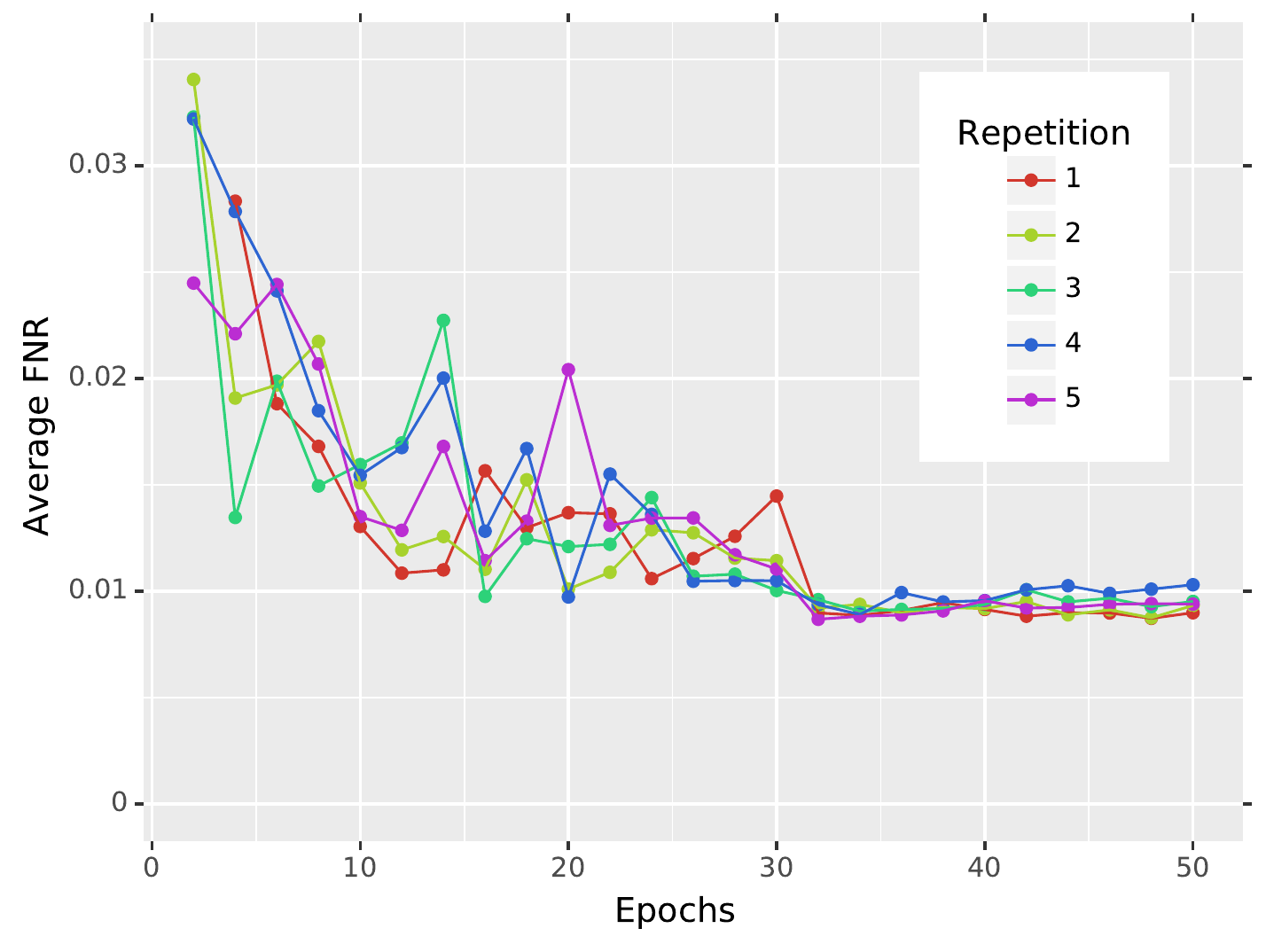} \\
			$\rho=0.1$ & $\rho=0.3$ & $\rho=0.5$
		\end{tabular}
		
	\end{center}
	\caption{ \label{fig:repetition} 
		Repeatability of the training procedure on BSD500 dataset.}
	
\end{figure*}

 In order to better understand the influence of different parameters on the final performance of the proposed network, we conducted some additional experiments. In the first one, we 
check whether the network impulse detection efficiency is repeatable when we start the training procedure from scratch. The changes of the average wACC, FPR,  and FNR calculated on test dataset during the training are presented in Fig. \ref{fig:repetition} and in Tab. \ref{tab:repetition}.

\begin{table}[]
	\caption[example] 
	{ \label{tab:repetition} 
		Repeatability of the training procedure on BSD500 dataset.}
	\begin{center}
		\begin{tabular}{clllll}
			\hline \hline 
			\multicolumn{6}{c}{Training repetition} \\ \cline{2-6}
			\multicolumn{6}{c}{Average wACC}                             \\ \cline{2-6}
			$\rho$ & 1      & 2      & 3      & 4      & 5      \\ \cline{2-6}
			0.1    & 0.9985 & 0.9987 & 0.9987 & 0.9985 & 0.9987 \\
			0.2    & 0.9984 & 0.9986 & 0.9986 & 0.9985 & 0.9985 \\
			0.3    & 0.9980 & 0.9981 & 0.9982 & 0.9981 & 0.9980 \\
			0.4    & 0.9970 & 0.9971 & 0.9972 & 0.9970 & 0.9970 \\
			0.5    & 0.9945 & 0.9944 & 0.9944 & 0.9941 & 0.9944 \\ \cline{2-6}
			\multicolumn{6}{c}{Average FPR}                             \\ \cline{2-6}
			$\rho$ & 1      & 2      & 3      & 4      & 5      \\ \cline{2-6}
			0.1    & 0.0014 & 0.0013 & 0.0013 & 0.0015 & 0.0012 \\
			0.2    & 0.0014 & 0.0012 & 0.0012 & 0.0014 & 0.0013 \\
			0.3    & 0.0014 & 0.0012 & 0.0012 & 0.0013 & 0.0013 \\
			0.4    & 0.0017 & 0.0014 & 0.0014 & 0.0013 & 0.0015 \\
			0.5    & 0.0020 & 0.0018 & 0.0016 & 0.0015 & 0.0018 \\ \cline{2-6}
			\multicolumn{6}{c}{Average FNR}                             \\ \cline{2-6}
			$\rho$ & 1      & 2      & 3      & 4      & 5      \\ \cline{2-6}
			0.1    & 0.0018 & 0.0016 & 0.0015 & 0.0015 & 0.0018 \\
			0.2    & 0.0025 & 0.0023 & 0.0022 & 0.0023 & 0.0024 \\
			0.3    & 0.0035 & 0.0034 & 0.0033 & 0.0034 & 0.0034 \\
			0.4    & 0.0051 & 0.0052 & 0.0050 & 0.0054 & 0.0051 \\
			0.5    & 0.0090 & 0.0093 & 0.0095 & 0.0103 & 0.0094 \\ \hline
		\end{tabular}
	\end{center}
	
\end{table}

As can be observed, the results are repeatable and independent from the experiment repetition and the networks start to stabilize and converge to its final performance when the learning rate 
is decreased after 30 epochs. Additionally, we can see that the average FPR is relatively low and wACC quite well reflects the network performance. Therefore, in the rest of the paper, we present the wACC metric only.

\par
\begin{figure} [ht]
	\begin{center}
		\begin{tabular}{c} 
			\includegraphics[width=0.45\textwidth]{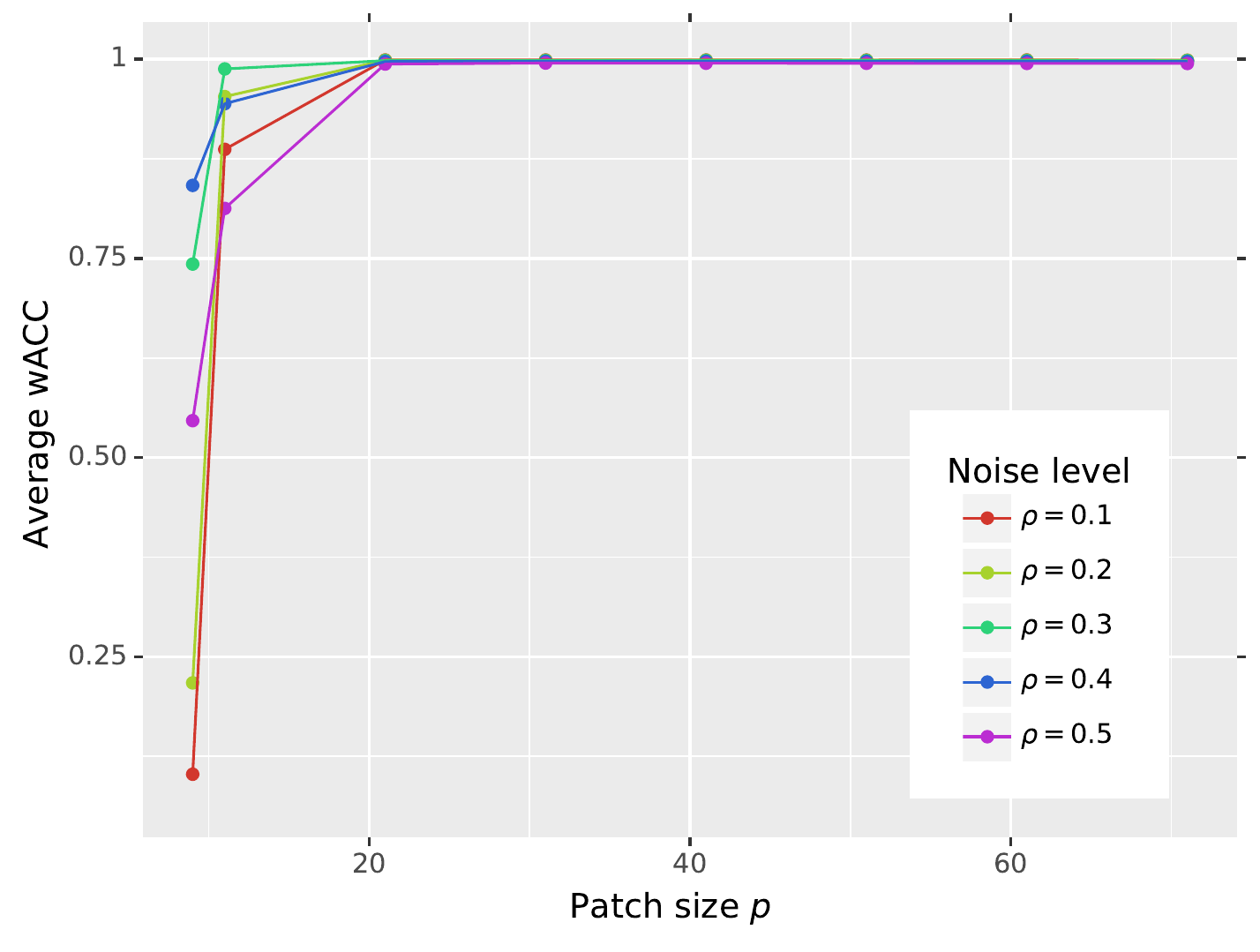} 
		\end{tabular}
		
	\end{center}
	\caption[example] 
	{ \label{fig:patch_size} 
		Impact of the patch size used in the training procedure on final detection weighted accuracy of the network.}
	
\end{figure}

 In the next experiment, we evaluated the influence of the patch size $p$ used in the training procedure on the final average performance of the network  (see  Fig. \ref{fig:patch_size} and Tab \ref{tab:patch_size}). We evaluated our method using the following patch sizes: $\{9,11,21,31,41,51,61,71 \}$. For smaller patch sizes, the network was not able to learn and therefore the results are not shown. 
\par As can be observed,  if the patch size used in the training is greater than or equal to $21 \! \times \! 21$, the optimal performance of the network is obtained. 
However, the increase of the patch size does not boost the network's performance, but it makes training more time consuming, because the loss function is calculated between bigger patches. Therefore, the selected patch size cannot be too small nor too big it would only increase the training time.

\begin{table}[]\scriptsize
	\caption[example] 
	{ \label{tab:patch_size} 
		Impact of the patch size used in the training procedure on final detection performance of the network.}
	\begin{center}
		\begin{tabular}{llllllll}
			\hline
			\multicolumn{8}{c}{Average wACC} \\ \hline
			& \multicolumn{7}{c}{Patch size $p$} \\ \cline{2-8}
			$\rho$ & 9      & 11     & 21     & 31     & 41     & 51     & 61      \\
			0.1    & 0.1005 & 0.1026 & 0.9987 & 0.9986 & 0.9987 & 0.9985 & {\bf 0.9988} \\
			0.2    & 0.2063 & 0.2159 & 0.9985 & 0.9985 & 0.9985 & 0.9985 & {\bf 0.9986} \\
			0.3    & 0.5258 & 0.5887 & 0.9980 & 0.9980 & {\bf 0.9981} & 0.9980 & 0.9980 \\
			0.4    & 0.6003 & 0.6107 & 0.9969 & {\bf 0.9972} &  0.9971 & 0.9970 & 0.9969 \\
			0.5    & 0.5000 & 0.5005 & 0.9938 & {\bf 0.9950} & 0.9946 & 0.9944 & 0.9942 \\ \hline
		\end{tabular}
	\end{center}
	
\end{table}

%

\begin{figure*} [ht]
	\begin{center}
		\begin{tabular}{c} 
			\includegraphics[width=0.92\textwidth]{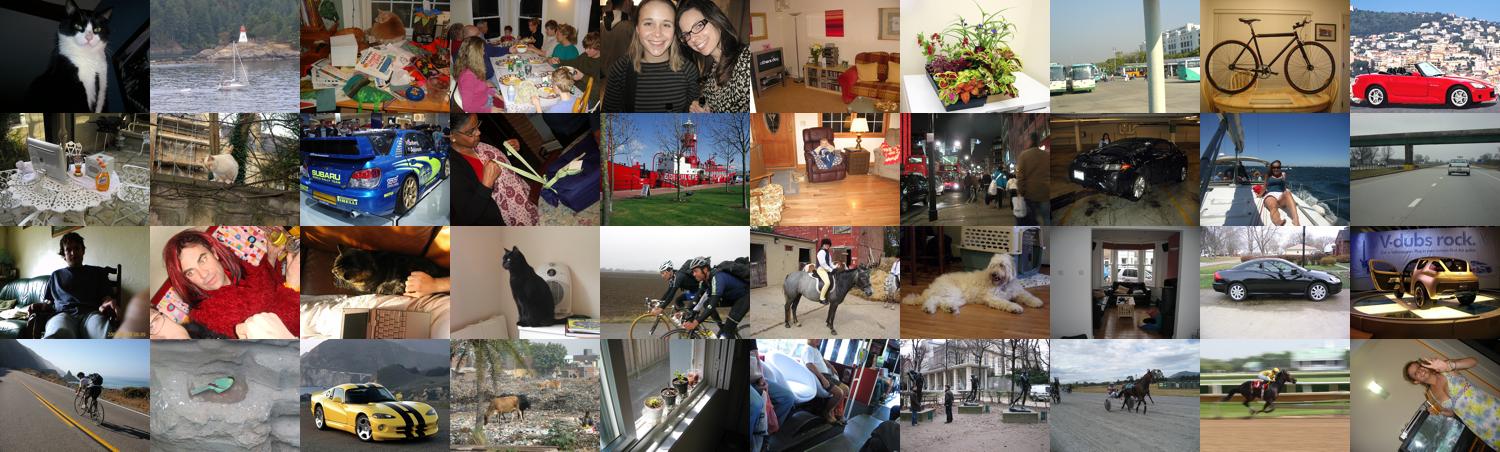}
		\end{tabular}
	\end{center}
	\caption
	{ \label{fig:pascal} 
		Example images from PASCAl VOC2007 \cite{pascal-voc-2007}.}
	
\end{figure*}

\begin{figure*} [ht]
	\begin{center}
		\begin{tabular}{c} 
			\includegraphics[width=0.92\textwidth]{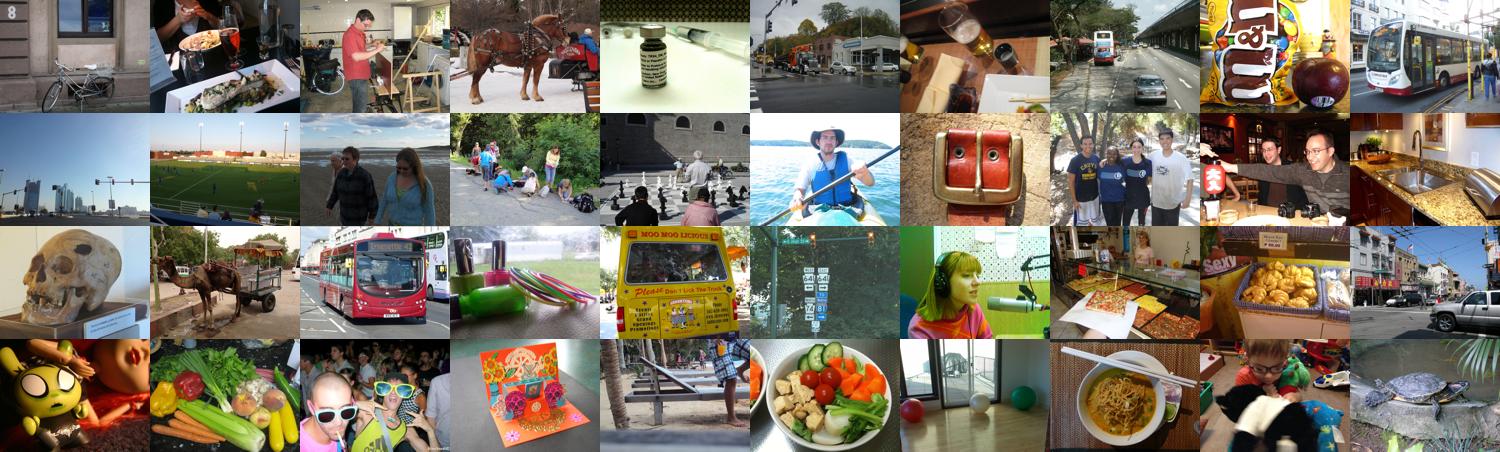}
		\end{tabular}
	\end{center}
	\caption 
	{ \label{fig:google} 
		Example images from GoogleV4 dataset \cite{OpenImages
		}.}
	
\end{figure*}

In the next experiment, we analyzed the impact of  the type of dataset used in training procedure and its size on the final network performance.  We selected two additional datasets: the PASAL VOC2007 dataset \cite{pascal-voc-2007} and the Google Open Images Dataset V4 (GoogleV4) \cite{OpenImages} introduced for object detection purposes. Both datasets contain high quality images and we selected randomly 500 pictures. The PASCAL VOC2007 dataset consists of images which present 20 classes of various objects. The GoogleV4 dataset contains images, which cover 600 classes, but in our research  we selected only 50 classes. Therefore both datasets consist of limited texture examples (e.g. images that depict only trains) since the content of the used test dataset is much wider. Additionally, for GoogleV4 dataset we also decreased the  original resolution four times to ensure similar resolution in all training and test datasets. 
Example images from both datasets are presented in Figs. \ref{fig:pascal} and \ref{fig:google}.

\begin{figure*} [ht]
	\begin{center}
		\begin{tabular}{ccc} 
			\includegraphics[width=0.31\textwidth]{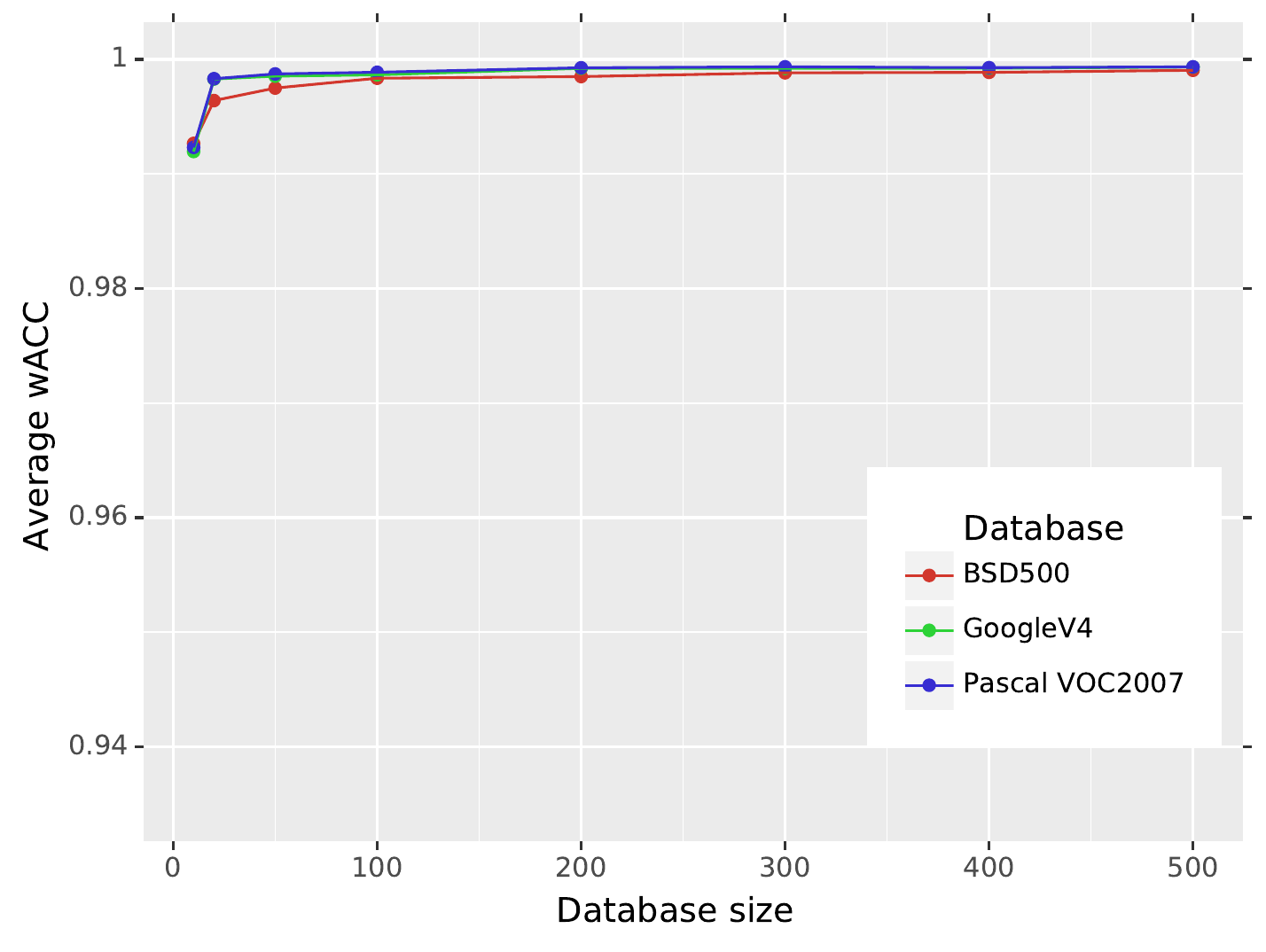} &
			\includegraphics[width=0.31\textwidth]{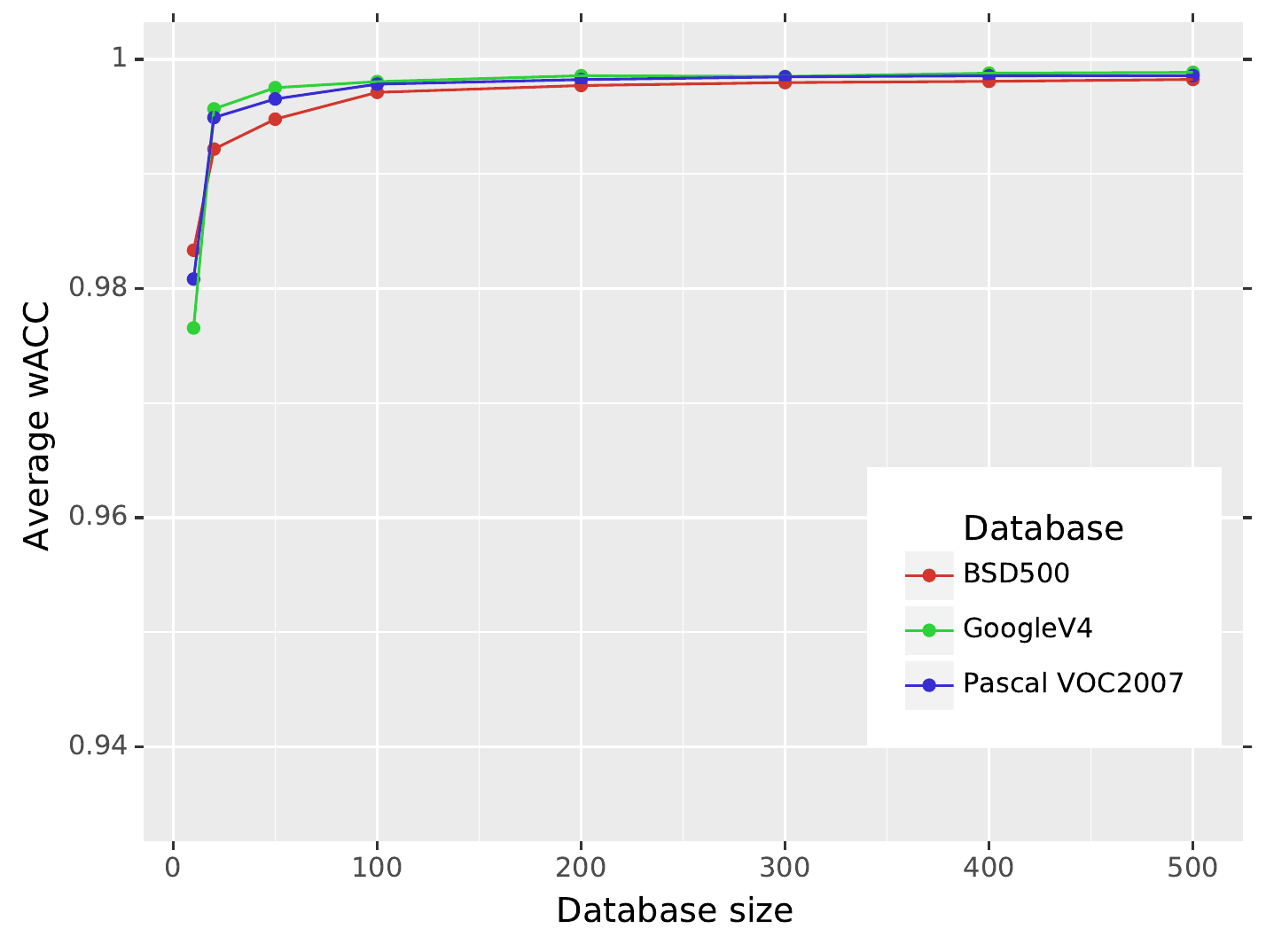} &
			\includegraphics[width=0.31\textwidth]{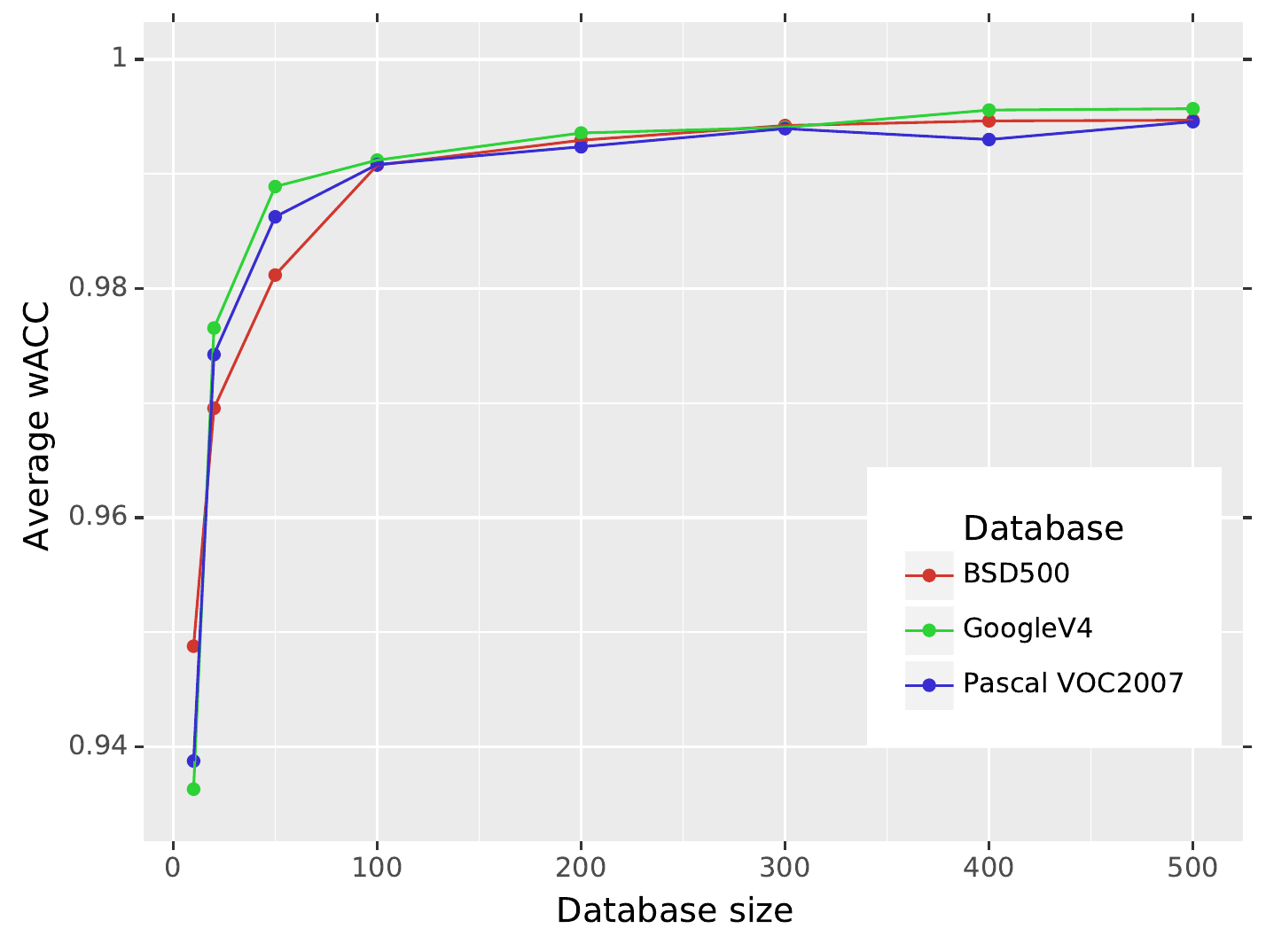} \\
			$\rho=0.1$ & $\rho=0.3$ & $\rho=0.5$
		\end{tabular}
		
	\end{center}
	\caption
	{ \label{fig:size} 
		Impact of the type of dataset used in the training and its size on the average wACC of the network.}
	
\end{figure*}

The influence of the size of the training dataset on the final average performance of the network is shown in  Fig. \ref{fig:size} and summarized in Tab. \ref{tab:size}. As can be observed, if the size of the dataset is increased, then the average wACC is also growing. The highest average wACC was achieved for GoogleV4 dataset, but the difference between various datasets is rather small. The performed experiment also shows that the training of the network requires sufficient amount of data in the training process to achieve expected effectiveness. However, the optimal performance can be achieved on different datasets.  Additionally, we can notice that 
when the noise density increases, the network needs more data in the training. Finally, we recommend to use 500 images in the training, which are divided into small patches in the training, but we need to remember that the final number of patches can differ depending on image resolution.  In one our experiment, the total number of non-overlapping patches of size $41 \! \times \! 41$ generated for BSD500 dataset was equal to 120500. 

\begin{table}[]\scriptsize
	\caption[example] 
	{ \label{tab:size} 
		Impact of the type of dataset used in the training and its size on the average wACC of the network.}
	\begin{center}
		\begin{tabular}{llllllll}
			\hline
			\multicolumn{8}{c}{Average wACC}		\\ \hline
			\multicolumn{8}{c}{BSD500}                                                   \\ \cline{1-8}
			& \multicolumn{7}{c}{Dataset size}                                      \\ \cline{2-8}
			$\rho$ & 10          & 50     & 100    & 200    & 300    & 400    & 500    \\ \cline{2-8}
			0.1 & 0.9910  & 0.9969 & 0.9983 & 0.9982 & 0.9985 & 0.9985 & {\bf 0.9987} \\
			0.2 & 0.9889  & 0.9962 & 0.9979 & 0.9982 & 0.9983 & 0.9984 & {\bf 0.9986} \\
			0.3 & 0.9830  & 0.9946 & 0.9970 & 0.9976 & 0.9978 & 0.9980 & {\bf 0.9982} \\
			0.4 & 0.9701  & 0.9909 & 0.9952 & 0.9962 & 0.9967 & 0.9970 & {\bf 0.9972} \\
			0.5 & 0.9434  & 0.9807 & 0.9898 & 0.9924 & 0.9938 & {\bf 0.9945} & 0.9944 \\ \hline
			\multicolumn{8}{c}{VOC2007}                                                 \\ \cline{1-8}
			& \multicolumn{7}{c}{Dataset size}                                      \\ \cline{2-8}
			$\rho$& 10         & 50     & 100    & 200    & 300    & 400    & 500    \\ \cline{2-8}
			0.1 & 0.9906  & 0.9985 & 0.9985 & 0.9992 & 0.9993 & 0.9991 & {\bf 0.9992} \\
			0.2 & 0.9872  & 0.9977 & 0.9985 & 0.9989 & 0.9990 & {\bf 0.9990} & {\bf 0.9990} \\
			0.3 & 0.9798  & 0.9962 & 0.9977 & 0.9982 & 0.9984 & 0.9985 & {\bf 0.9986} \\
			0.4 & 0.9647  & 0.9931 & 0.9958 & 0.9967 & 0.9972 & 0.9973 & {\bf 0.9974} \\
			0.5 & 0.9345  & 0.9849 & 0.9894 & 0.9922 & 0.9930 & 0.9926 & {\bf 0.9939} \\ \hline
			\multicolumn{8}{c}{GoogleV4 dataset}                                   \\ \cline{1-8}
			& \multicolumn{7}{c}{Dataset size}                                      \\ \cline{2-8}
			$\rho$ & 10         & 50     & 100    & 200    & 300    & 400    & 500    \\ \cline{2-8}
			0.1 & 0.9907  & 0.9980 & 0.9983 & 0.9990 & 0.9986 & 0.9990 & {\bf 0.9993} \\
			0.2 & 0.9850  & 0.9979 & 0.9983 & 0.9988 & 0.9987 & 0.9990 & {\bf 0.9991} \\
			0.3 & 0.9766  & 0.9973 & 0.9979 & 0.9985 & 0.9985 & 0.9987 & {\bf 0.9988} \\
			0.4 & 0.9614  & 0.9954 & 0.9964 & 0.9975 & 0.9975 & 0.9980 & {\bf 0.9981} \\
			0.5 & 0.9332  & 0.9886 & 0.9900 & 0.9932 & 0.9937 & 0.9950 & {\bf 0.9952}\\
			\hline
		\end{tabular}
	\end{center}
	
\end{table}

\begin{figure} [ht]
	\begin{center}
		\includegraphics[width=0.48\textwidth]{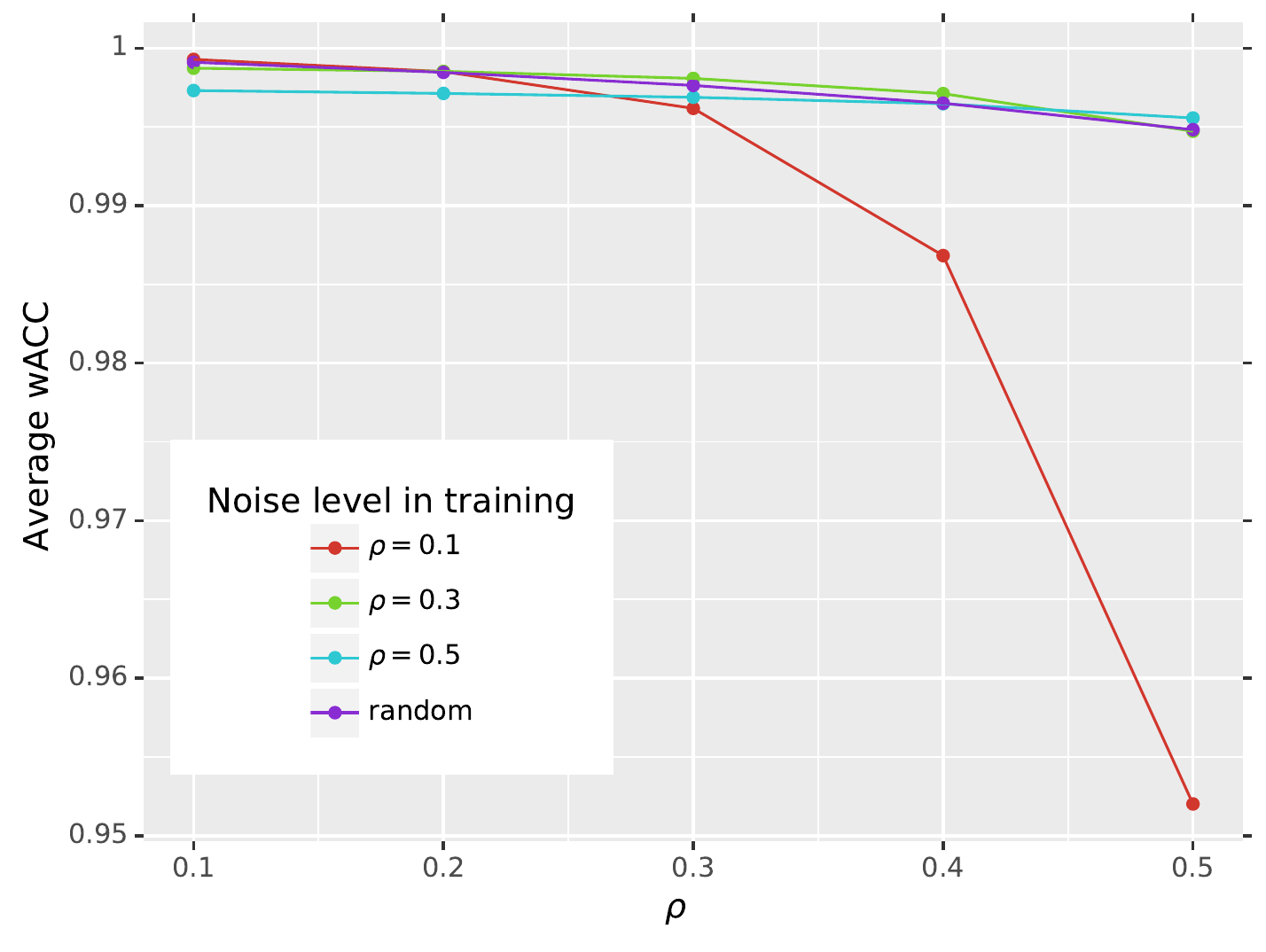}

	\end{center}
	\caption[example] 
	{ \label{fig:noise_lvl} 
		Impact of the noise density used during training  on the final network performance and its ability to detect impulses in test images contaminated with different noise density.}
	
\end{figure}

The last issue that we would like to address in the scope of this work was what noise density level should be used in the training procedure to obtain optimal network performance. For the original DnCNN, the authors proved that it was not so important what intensity level of Gaussian noise was used in the training and the network trained with $\sigma=25$ was able to denoise images corrupted with different noise density. To confirm this behavior for impulsive noise, we trained the network with patches contaminated with noise density $\rho=\{0.1,0.3,0.5 \}$. Additionally, we trained the network with patches contaminated with randomly selected  noise probability 
from the range $[0.1,0.5]$.
This experiment is denoted in this paper as random and the results are depicted in Fig \ref{fig:noise_lvl} and summarized in Tab. \ref{tab:noise_lvl}.

\begin{table}[]
	\caption[example] 
	{ \label{tab:noise_lvl} 
		Impact of the noise density used during training  on the final network performance and its ability to detect impulses in test images contaminated with different noise density.}
	\begin{center}
		\begin{tabular}{lllll}
			\hline
			\multicolumn{5}{c}{Average wACC} \\ \hline
			& \multicolumn{4}{l}{Noise density in the training} \\ \cline{2-5}
			$\rho$& random           & 0.1           & 0.3          & 0.5          \\ \cline{2-5}
			0.1 & 0.9991        & {\bf 0.9993}      & 0.9987       & 0.9973       \\
			0.2 & {\bf 0.9985}       &  {\bf 0.9985}        & {\bf 0.9985}      & 0.9971       \\
			0.3 & 0.9976        & 0.9962        & {\bf 0.9981}      & 0.9969       \\
			0.4 & 0.9965        & 0.9868        & {\bf 0.9971}       & 0.9965       \\
			0.5 & 0.9948        & 0.9520        & 0.9947       & {\bf 0.9956} \\ \hline     
		\end{tabular}
	\end{center}
	
\end{table}

As can be observed, the highest values are observed if the noise level 
during training and during tests was the same. In case of using a random noise density during  training, the results are very close to the optimal performance. The highest deviations from the optimal average wACC were obtained if the patches were contaminated with low noise density level during training and heavy noise at testing phase and vice versa. In most cases, the optimal wACC was obtained using $\rho=0.3$ and for other noise levels in the test phase, the performance was very close to the optimal, therefore we recommend to use this value 
during training. 
\par To summarize, the performed experiments confirmed that the analyzed parameters proposed for DnCNN can be used as a default for the proposed IDCNN. However, it will be indispensable in the future to more carefully analyze the impact of the diversity of the dataset used 
during training on the final performance of the network.

\section{ Comparison with the state-of-the-art denoising
	methods}

The proposed switching filter that can be regarded as an extension of the DnCNN to detect impulses in the image with the adaptive mean filter for impulses restoration was compared with state-of-the-art methods using quantitative measures such as the Peak Signal to Noise Ratio (PSNR) and the Mean Absolute Error (MAE).

The PSNR measure is defined as
\begin{flalign}
&\textrm{PSNR}=10\cdot \log _{{10}}\left({\frac  {{  {255}}^{2}}{{\textrm{MSE}}}}\right),\\
&\textrm{MSE}={\frac  {1}{3Q}}\sum _{{i=1}}^{{Q}}\sum _{{q=1}}^{{3}}[x_{i,q} -\hat{x}_{i,q}]^{2}, \\
\end{flalign}
where each $x_{i,q}, q=1,2,3$ denotes  the pixels of the original image, $ \hat{x}_{i,q}, q = 1,2,3$ denotes the pixels of the restored image and $Q$ denotes the number of pixels in an image.
The MAE metric is defined as
\begin{flalign}
&\textrm{MAE}={\frac  {1}{3Q}}\sum _{{i=1}}^{{Q}}\sum _{{q=1}}^{{3}}|x_{i,q} -\hat{x}_{i,q}|.
\end{flalign}
Additionally, we used the Structural SIMilarity index (SSIM\textsubscript{c}) designed for color images \cite{WangBovik2004}, because it has demonstrated better agreement with human observers in image quality assessment than traditional metrics.

\par In this work, the following state-of-the-art filters were taken for comparison:  Denoising Convolutional Neural Network trained on impulsive noise (DnCNN) \cite{RadlakMalinski2019}, Fast Averaging Peer Group Filter (FAPGF) \cite{Mal1}, Fast Adaptive Switching Trimmed Arithmetic Mean Filter (FASTAMF) \cite{Mal2}, Fast Fuzzy Noise Reduction Filter (FFNRF) \cite{Mor5},  Fuzzy  Rank-Ordered Differences Filter (FRF) \cite{CAMARENA2010},
Impulse Noise Reduction Filter (INRF) \cite{SchulteMorillas2007}, Patch-based Approach for the Restoration of Images affected by Gaussian and Impulse noise (PARIGI) \cite{Delon2013},  Peer Group Filter (PGF) \cite{KenneyDeng2001}.  As our final method in evaluation, we used two switching filters. Both methods of impulse detection used proposed IDCNN, but the final networks were trained on two different datasets: BSD500 (IDCNN\textsubscript{BSD500}) and GoogleV4 (IDCNN\textsubscript{Google}). 
\par
\begin{figure}[ht]\centering
	\begin{tabular}{cc}
		
		\includegraphics[width=0.22\textwidth]{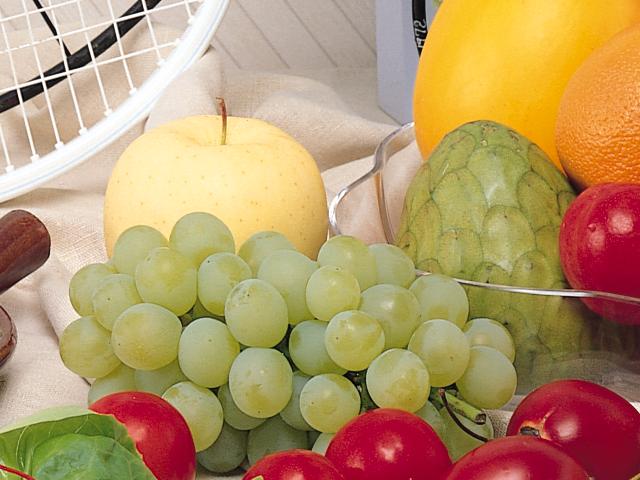} &
		\includegraphics[width=0.22\textwidth]{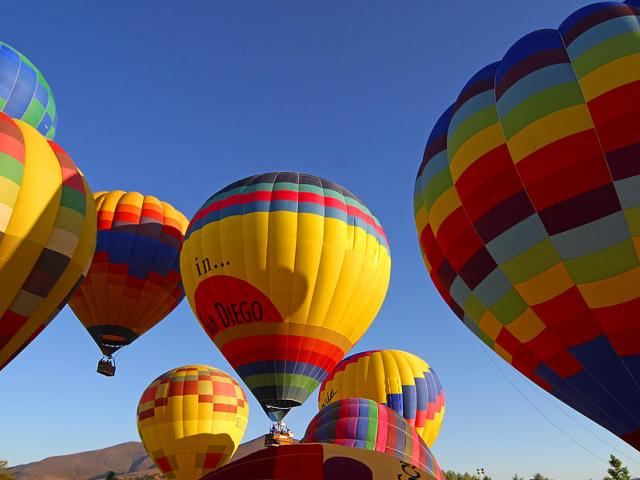} \\
		(a) FRUITS & (b) BALLOONS  \\ 
		\includegraphics[width=0.22\textwidth]{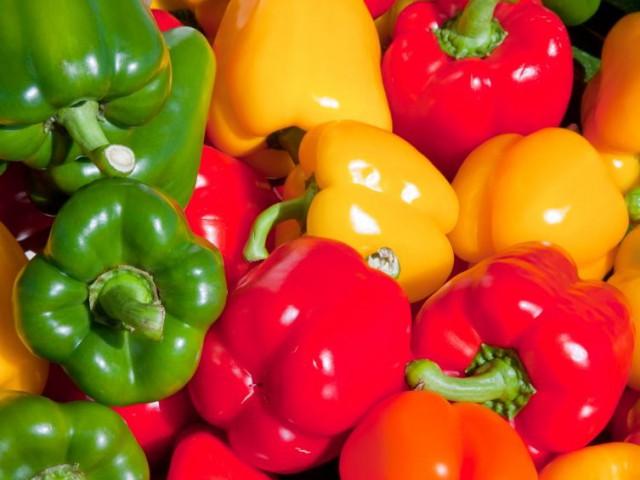} &
		\includegraphics[width=0.22\textwidth]{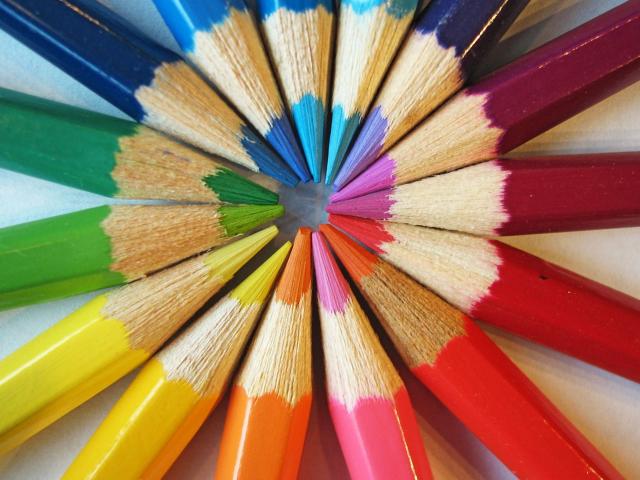} \\
		(c) PEPPERS & (d) CRAYONS
		
	\end{tabular}
	\medskip
	\caption{Representative test images from benchmark dataset \cite{Mal1} for which numerical results were calculated. \label{fig:four_test_images}}
\end{figure}

 The numerical results are shown in Table \ref{tab:dncnn_psnr} using four representative test images chosen from the test dataset \cite{Mal1}, which are presented in Fig. \ref{fig:four_test_images}. As can be observed,  in all cases, the results of the proposed switching filter that uses IDCNN for impulse detection and adaptive mean filter for image restoration outperform other state-of-the-art techniques.

\par
\begin{figure*}[!h]\centering
	\begin{tabular}{ccc}
		
		\includegraphics[width=0.31\textwidth]{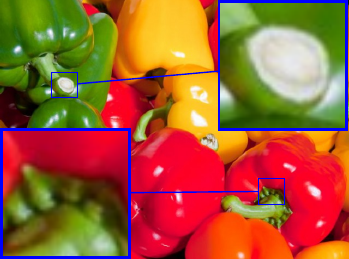} &
		\includegraphics[width=0.31\textwidth]{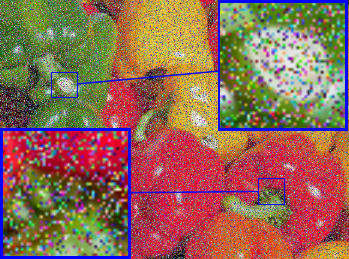} &
		\includegraphics[width=0.31\textwidth]{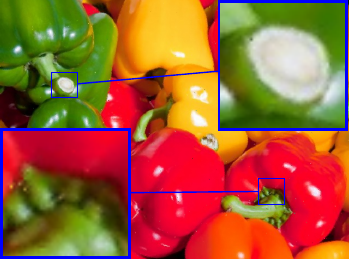} \\
		
		(a) ORIGINAL & (b) NOISY, $\rho$=0.4 & (c) IDCNN\textsubscript{BSD500} \\

		\includegraphics[width=0.31\textwidth]{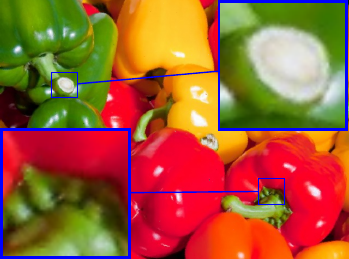} &
		\includegraphics[width=0.31\textwidth]{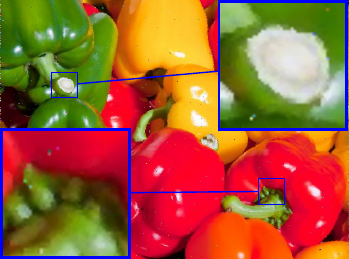} &
		\includegraphics[width=0.31\textwidth]{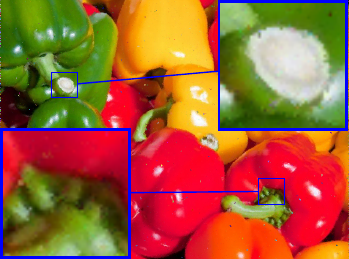} \\
		
		(d) IDCNN\textsubscript{GoogleV4}& (e) FRF & (f) INRF \\
		
		\includegraphics[width=0.31\textwidth]{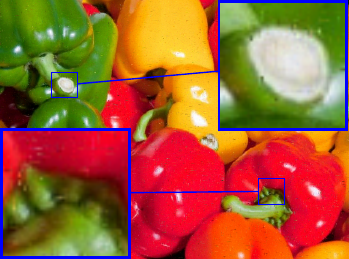} &
		\includegraphics[width=0.31\textwidth]{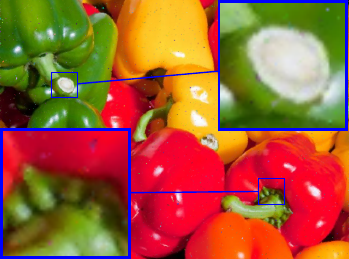} &
		\includegraphics[width=0.31\textwidth]{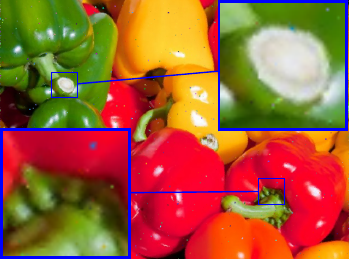} \\
		(g) DnCNN &	(h) FAPGF & (i) FASTAMF  \\

		\includegraphics[width=0.31\textwidth]{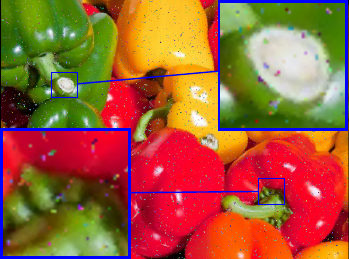} &
		\includegraphics[width=0.31\textwidth]{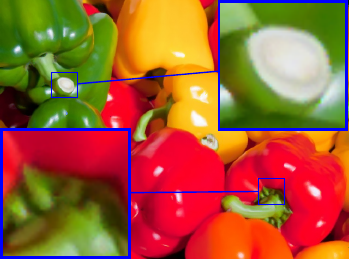} &
		\includegraphics[width=0.31\textwidth]{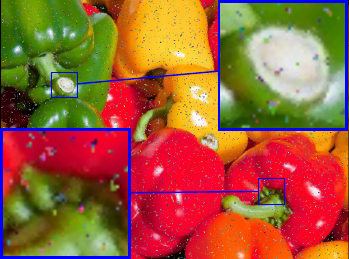} \\
		(g) FFNRF  &	(h) PARIGI   & (i) PGF 	
	\end{tabular}
	\caption{Visual comparison of the filtering efficiency using a part of the PEPPERS image ($\rho$ = 0.4). \label{fig:motoross}}
\end{figure*}

 The visual comparison of the obtained results is depicted in Fig.   \ref{fig:motoross}.  As can be observed, the proposed IDCNN  is able to correctly localize almost all impulses and the visible artifacts are the effect of insufficient quality of restoration of detected impulses. Therefore, future work will be focused on the improvement of the efficiency of the noisy pixel replacement method.
\par
\begin{figure*}[!h]\centering
	\begin{tabular}{ccc}
		
		\includegraphics[width=0.31\textwidth]{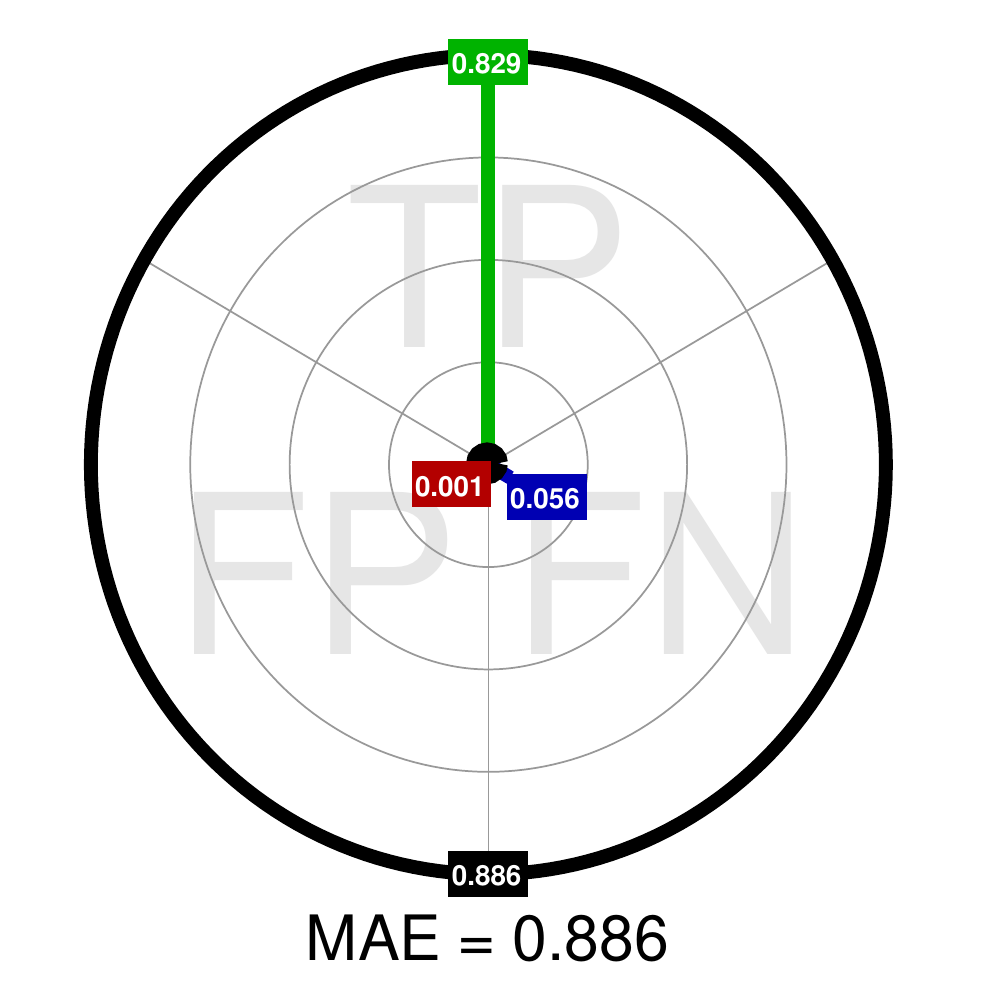} &
		\includegraphics[width=0.31\textwidth]{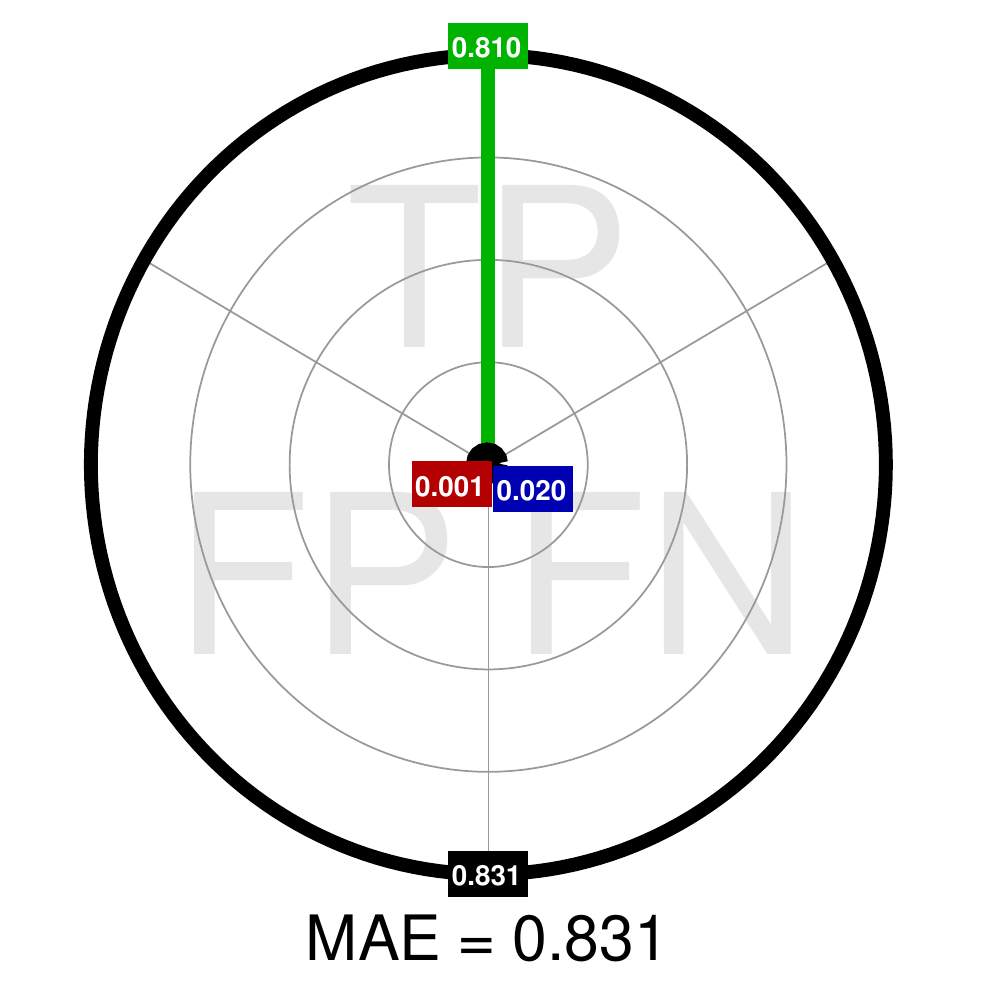}  & \includegraphics[width=0.31\textwidth]{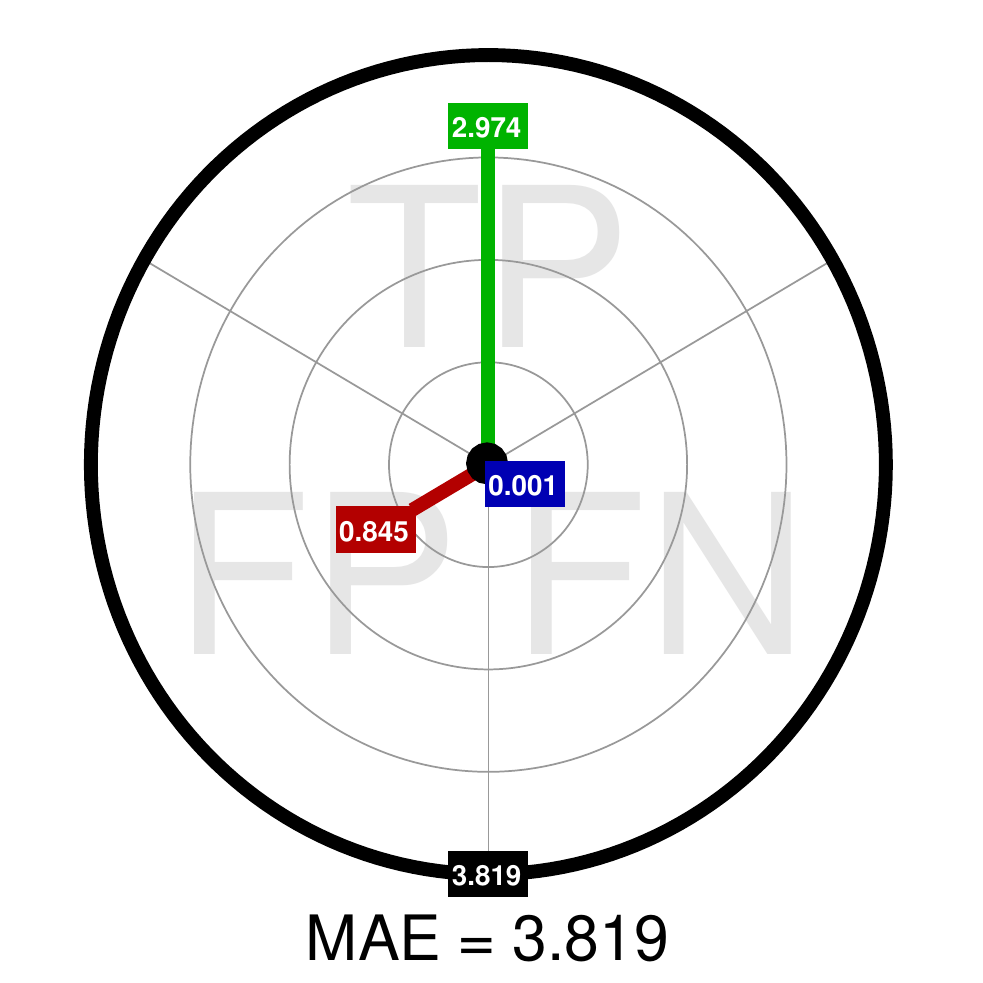}
		\\ 
		
		(a) IDCNN\textsubscript{BSD500}& (b) IDCNN\textsubscript{GoogleV4}& (c) DnCNN\\

		\includegraphics[width=0.31\textwidth]{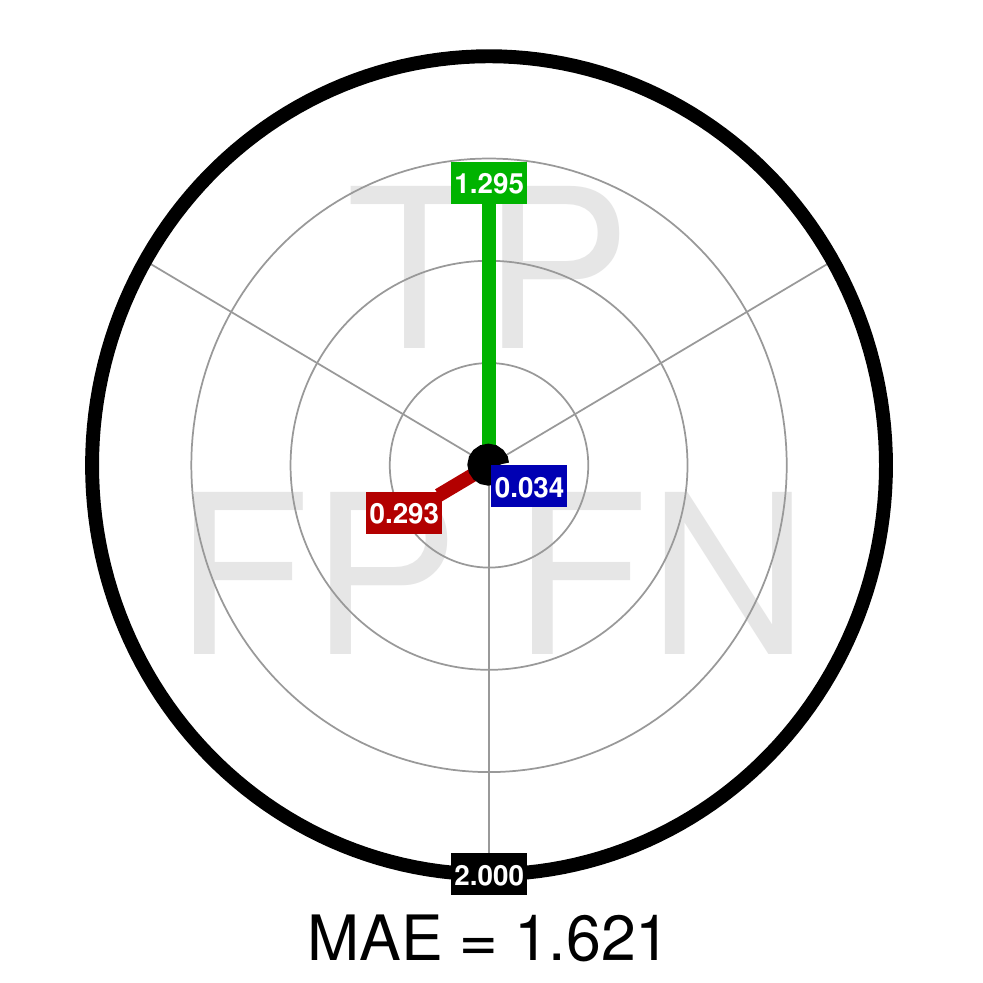}  &
		\includegraphics[width=0.31\textwidth]{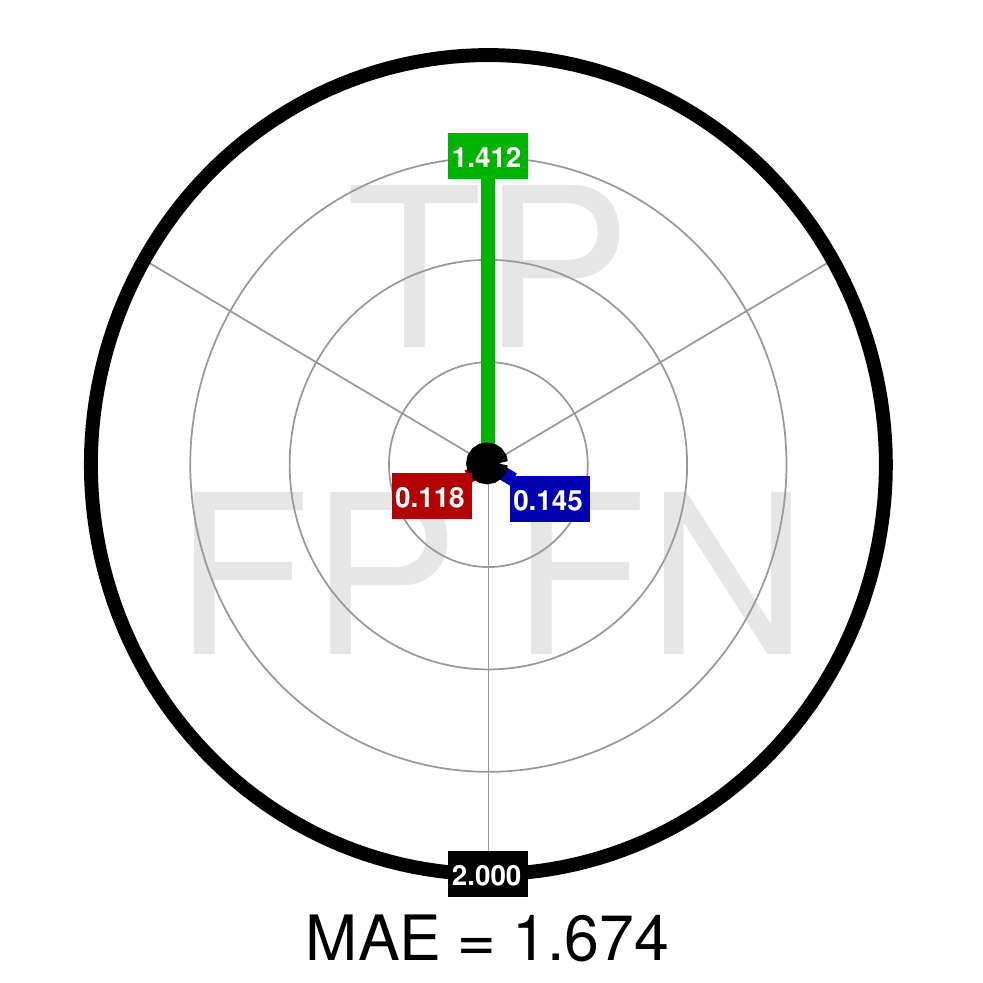}  &
		\includegraphics[width=0.31\textwidth]{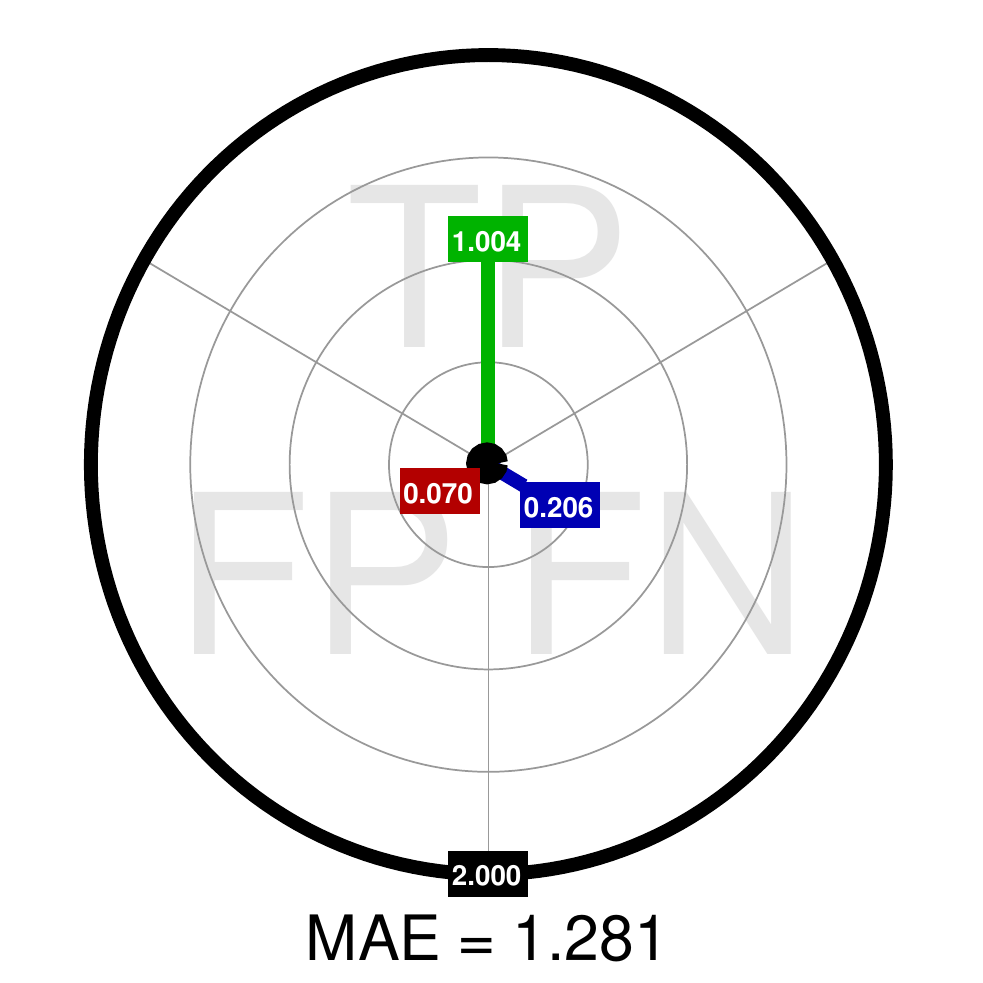}  \\
		
		(g) FRF  &	(h) FAPGF & (i) FASTAMF \\
		
		\includegraphics[width=0.31\textwidth]{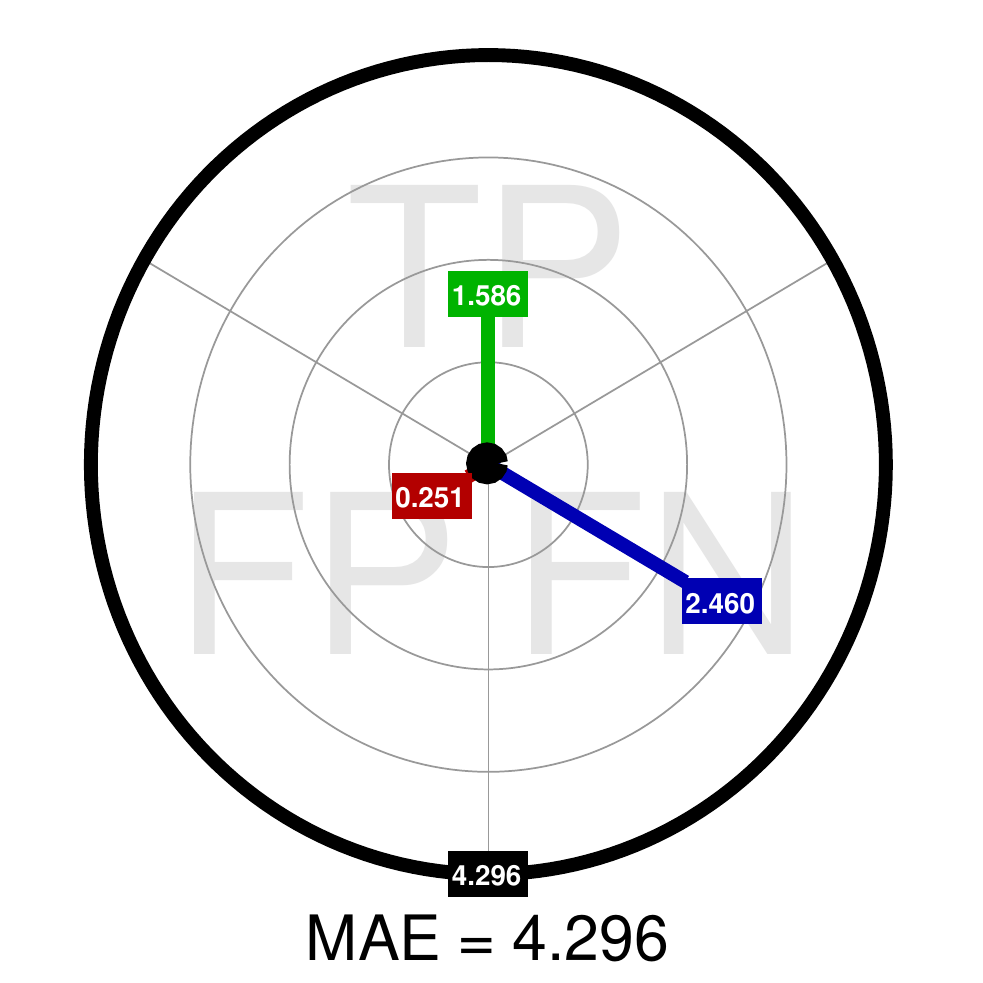} &
		\includegraphics[width=0.31\textwidth]{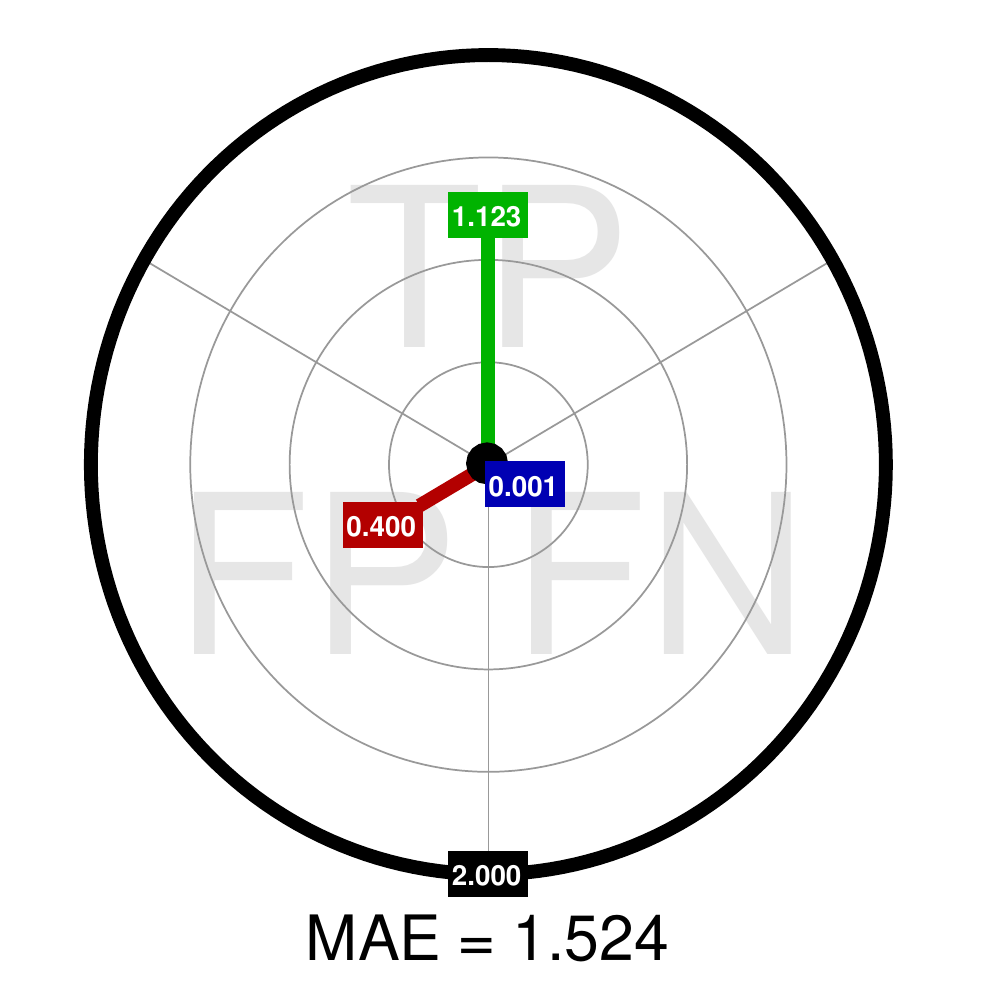}  &
		\includegraphics[width=0.31\textwidth]{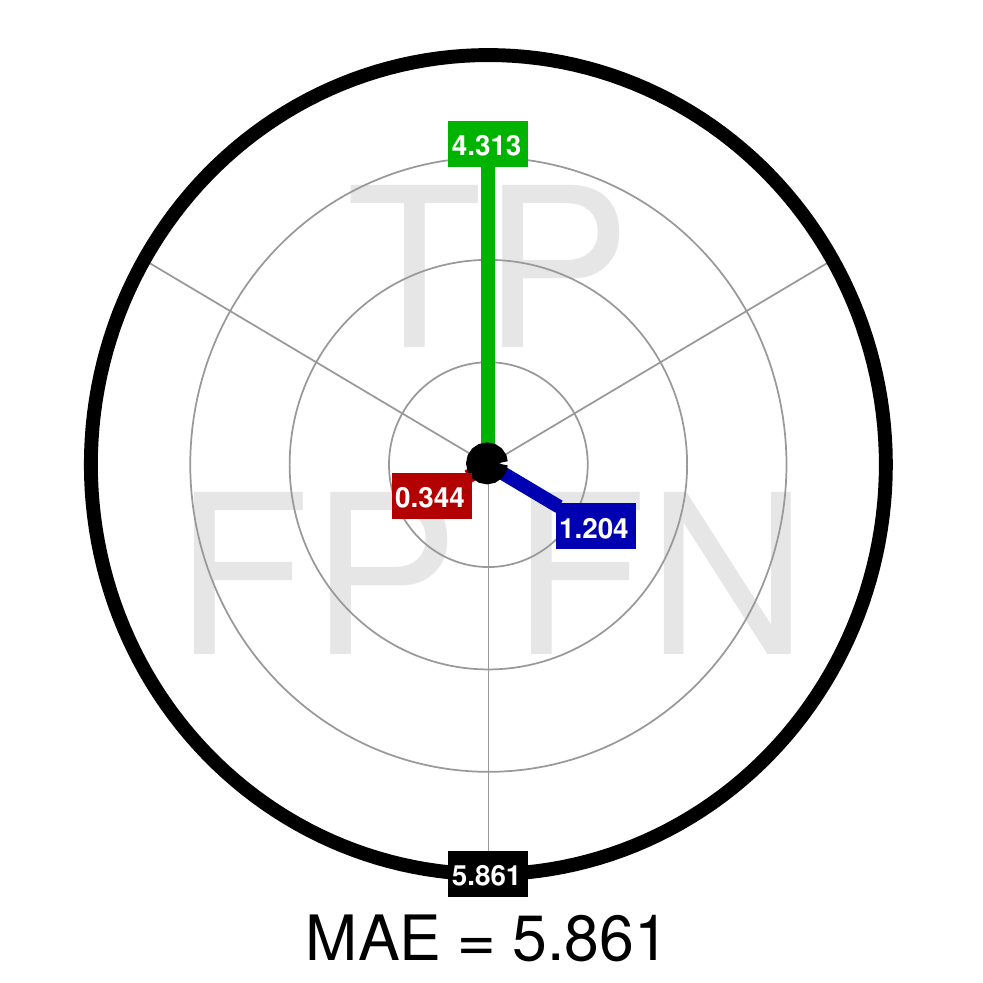}  \\
		
		(g) FFNRF  &	(h) PARIGI   & (i) PGF  
	\end{tabular}
	\caption{The diagrams that show what portion of the MAE error was caused by the improper decision of the used filter from classification perspective. These diagrams were obtained for the PEPPERS image ($\rho$ = 0.4). The main contribution to the total error is made by the replacement (interpolation) of the correctly detected impulses. \label{fig:motocross_circle}}
\end{figure*}

  To confirm that the main source of error is the method of interpolation, we presented the Aim Diagram (AD) which separates the distribution of errors that was caused by improper classification of the impulses. Using traditional metrics, we are not able to evaluate 
whether the main source of the error was caused by incorrect impulse detection or corrupted pixels restoration. In the proposed diagrams, the radiuses  in the circle denote the proportion of the MAE metric calculated independently for pixels that are TP, FP and FN respectively.  The error for TN pixels is equal to zero and therefore this radius is not presented on the plots. The AD calculated for MAE metric is presented in Fig. \ref{fig:motocross_circle}. As can be observed, the proposed impulse detector CNN\textsubscript{BSD500} and CNN\textsubscript{GoogleV4}, trained on BSD500 and GoogleV4 datasets respectively, almost perfectly detected all impulses and the main contribution to the MAE error comes from 
insufficient quality of detected impulse restoration. It shows that in further research the quality of the proposed filter could be improved if we would use a better noisy pixel interpolation method.

\par
\begin{figure*}[ht]\centering
	\begin{tabular}{c}
		$\rho=0.3$ \\
		\includegraphics[width=0.65\textwidth]{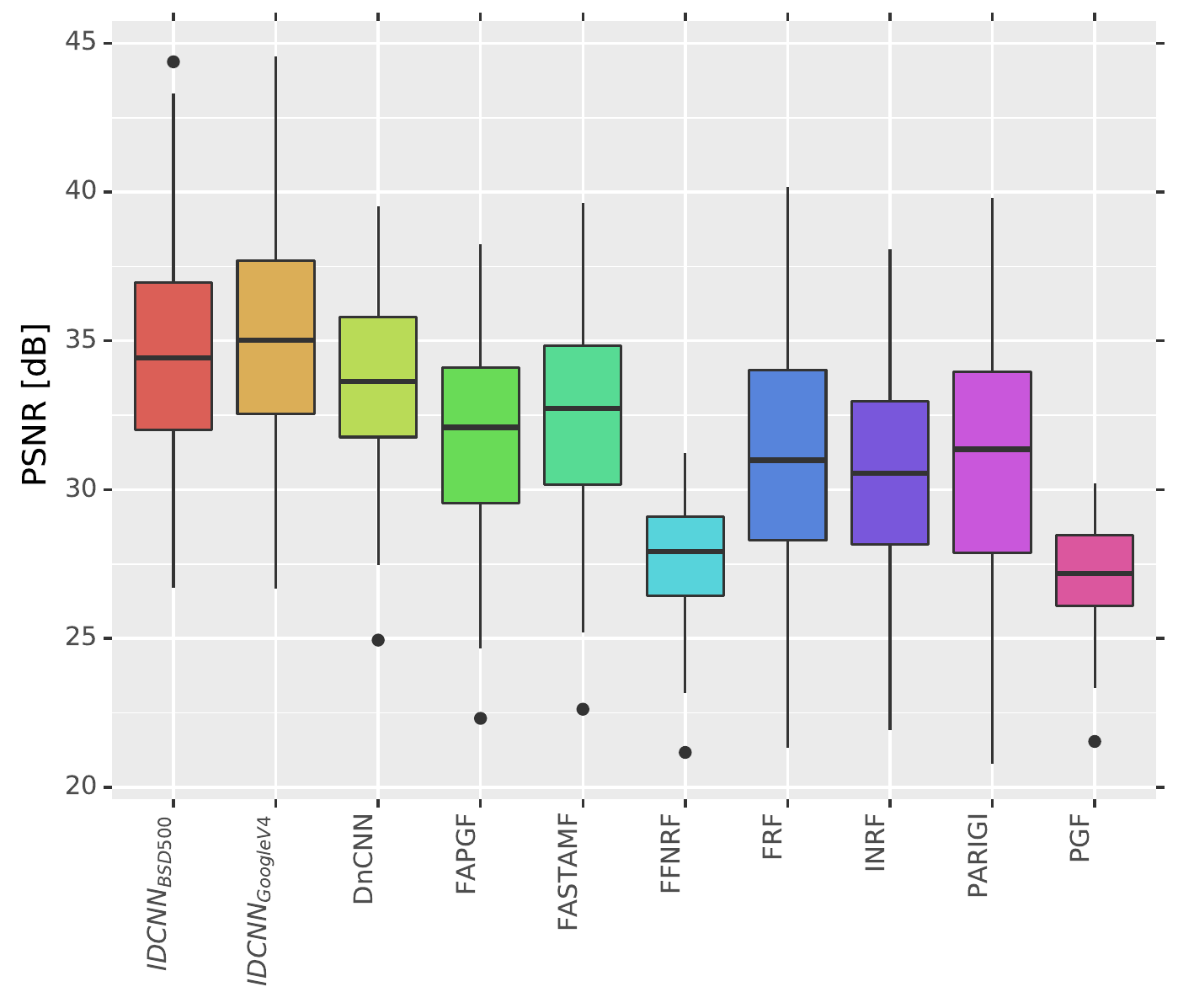}
		\\ 
		$\rho=0.5$ \\
		\includegraphics[width=0.65\textwidth]{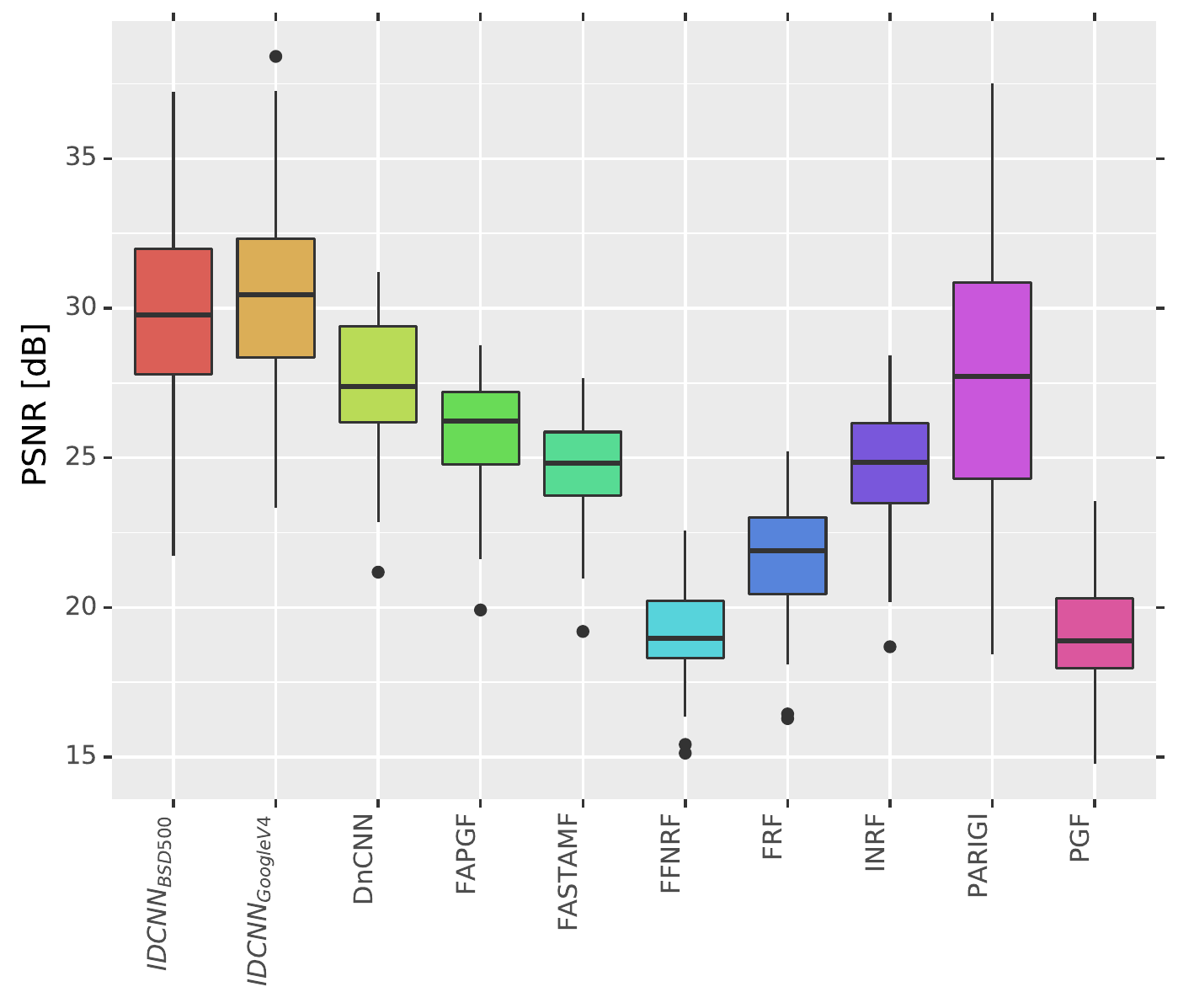}

	\end{tabular}
	\caption{The representative boxplots that shows the distribution of the obtained results for the analyzed methods on the test dataset \cite{Mal1}. \label{fig:ig:histograms_psnr}}
\end{figure*}

 Finally, the average values of selected metrics calculated on the test dataset \cite{Mal1} are presented in Tab. \ref{tab:test_final}. We also included the representative boxplots for PSNR measure to show the distribution of the obtained results (see Fig. \ref{fig:ig:histograms_psnr}). As can be observed, the average results of the proposed filter based on deep learning are significantly better than state-of-the-arts filters in terms of all used metrics. Additionally, the proposed switching filter allows achieving much better results than original DnCNN trained for impulsive noise.

\begin{table*}[] \scriptsize
	\caption{Comparison of the denoising efficiency of the proposed network for impulsive noise removal with the state-of-the-art methods on selected representative images from test dataset \cite{Mal1}. \label{tab:dncnn_psnr}}
	\setlength{\extrarowheight}{0pt}
	\begin{center}
		
		\begin{tabular}{lllllllllll} \hline \hline
			$\rho$ & IDCNN\textsubscript{BSD500} & IDCNN\textsubscript{GoogleV4} & DnCNN & FAPGF & FASTAMF & FFNRF & FRF   & INRF & PARIGI & PGF   \\ \toprule[2pt]
			\multicolumn{11}{c}{FRUITS}                                                                           \\ \toprule[2pt]
			\multicolumn{11}{c}{PSNR {[}dB{]}}                                                                    \\ \hline
			0.1    & 39.78       & \textbf{40.35 }      & 35.18 & 37.61 & 38.30   & 36.97 & 36.03 & 33.90 & 34.66  & 37.03 \\
			0.2    & 36.54       & \textbf{37.24}       & 36.47 & 34.69 & 35.47   & 33.20 & 33.21 & 32.03 & 33.14  & 32.52 \\
			0.3    & 34.22       & \textbf{35.00}       & 33.75 & 32.19 & 32.70   & 28.44 & 30.80 & 30.28 & 31.00  & 27.85 \\
			0.4    & 32.37       & \textbf{33.20 }      & 31.37 & 29.89 & 29.86   & 23.95 & 27.62 & 28.43 & 29.25  & 23.52 \\
			0.5    & 29.71       & \textbf{30.08}       & 27.87 & 26.53 & 25.40   & 19.88 & 22.27 & 25.26 & 26.81  & 19.67 \\ \hline
			\multicolumn{11}{c}{MAE}                                                                              \\ \hline
			0.1    & 0.45        & \textbf{0.42}        & 1.27  & 0.55  & 0.50    & 0.51  & 0.55  & 0.84  & 1.51   & 0.53  \\
			0.2    & 0.88        & \textbf{0.84 }       & 1.66  & 1.07  & 0.96    & 1.08  & 1.07  & 1.41  & 1.93   & 1.21  \\
			0.3    & 1.36        & \textbf{1.30}        & 2.31  & 1.73  & 1.52    & 2.12  & 1.72  & 2.14  & 2.51   & 2.40  \\
			0.4    & 1.90        & \textbf{1.81 }       & 3.25  & 2.61  & 2.30    & 4.16  & 2.72  & 3.14  & 3.10   & 4.69  \\
			0.5    & 2.70        & \textbf{2.65}        & 5.03  & 4.20  & 4.07    & 8.26  & 5.53  & 4.99  & 3.93   & 9.21  \\ \hline
			\multicolumn{11}{c}{SSIM}                                                                             \\ \hline
			0.1    & 0.985       & \textbf{0.987 }      & 0.972 & 0.979 & 0.983   & 0.978 & 0.978 & 0.973 & 0.935  & 0.974 \\
			0.2    & 0.970       & \textbf{0.974 }      & 0.953 & 0.956 & 0.966   & 0.940 & 0.958 & 0.947 & 0.918  & 0.927 \\
			0.3    & 0.950       & \textbf{0.958}       & 0.920 & 0.921 & 0.941   & 0.842 & 0.934 & 0.909 & 0.896  & 0.824 \\
			0.4    & 0.925       & \textbf{0.936 }      & 0.861 & 0.861 & 0.893   & 0.649 & 0.877 & 0.843 & 0.872  & 0.640 \\
			0.5    & 0.880       & \textbf{0.886 }      & 0.744 & 0.746 & 0.772   & 0.420 & 0.660 & 0.721 & 0.842  & 0.433 \\ \toprule[2pt]
			\multicolumn{11}{c}{BALLOONS}                                                                          \\ \toprule[2pt]
			\multicolumn{11}{c}{PSNR {[}dB{]}}                                                                    \\ \hline
			0.1    & 40.09       & \textbf{40.91}       & 36.65 & 36.94 & 38.18   & 37.09 & 36.26 & 34.52 & 36.15  & 37.27 \\
			0.2    & 36.83       & \textbf{37.75 }      & 34.52 & 34.56 & 35.54   & 33.35 & 33.16 & 32.84 & 34.74  & 32.59 \\
			0.3    & 34.17       & \textbf{35.13}       & 32.22 & 32.24 & 32.94   & 28.37 & 30.54 & 31.02 & 32.24  & 27.53 \\
			0.4    & 31.90       & \textbf{33.30}       & 30.03 & 29.90 & 30.04   & 23.52 & 27.29 & 28.75 & 30.13  & 22.90 \\
			0.5    & 28.56       & \textbf{29.85}       & 26.78 & 26.25 & 25.11   & 19.20 & 21.89 & 25.14 & 28.31  & 18.85 \\ \hline
			\multicolumn{11}{c}{MAE}                                                                              \\ \hline
			0.1    & 0.31        & \textbf{0.29}        & 1.18  & 0.45  & 0.37    & 0.35  & 0.41  & 0.67  & 0.81   & 0.38  \\
			0.2    & 0.61        & \textbf{0.58}        & 1.66  & 0.83  & 0.70    & 0.77  & 0.82  & 1.05  & 1.09   & 0.88  \\
			0.3    & 0.97        & \textbf{0.91}        & 2.31  & 1.37  & 1.11    & 1.65  & 1.36  & 1.62  & 1.53   & 1.98  \\
			0.4    & 1.40        & \textbf{1.26}        & 3.30  & 2.14  & 1.71    & 3.71  & 2.23  & 2.46  & 1.99   & 4.43  \\
			0.5    & 2.26        & \textbf{1.99}        & 5.39  & 3.77  & 3.49    & 8.29  & 4.88  & 4.34  & 2.58   & 9.69  \\ \hline
			\multicolumn{11}{c}{SSIM}                                                                             \\ \hline
			0.1    & 0.993       & \textbf{0.994}       & 0.978 & 0.983 & 0.989   & 0.987 & 0.985 & 0.982 & 0.976  & 0.980 \\
			0.2    & 0.985       & \textbf{0.986}       & 0.960 & 0.963 & 0.977   & 0.955 & 0.969 & 0.964 & 0.969  & 0.939 \\
			0.3    & 0.973       & \textbf{0.977}       & 0.926 & 0.929 & 0.960   & 0.851 & 0.948 & 0.932 & 0.955  & 0.823 \\
			0.4    & 0.954       & \textbf{0.966}       & 0.852 & 0.863 & 0.920   & 0.635 & 0.900 & 0.867 & 0.941  & 0.612 \\
			0.5    & 0.902       & \textbf{0.923}       & 0.690 & 0.724 & 0.783   & 0.378 & 0.690 & 0.725 & 0.920  & 0.375 \\ \toprule[2pt]
			\multicolumn{11}{c}{PEPPERS}                                                                          \\ \toprule[2pt]
			\multicolumn{11}{c}{PSNR {[}dB{]}}                                                                    \\ \hline
			0.1    & 47.75       & \textbf{47.88}       & 40.50 & 44.52 & 45.88   & 43.33 & 44.09 & 40.83 & 39.72  & 41.21 \\
			0.2    & 43.85       & \textbf{44.41 }      & 36.92 & 40.54 & 41.34   & 35.52 & 39.99 & 37.86 & 38.43  & 33.44 \\
			0.3    & 40.67       & \textbf{41.64}       & 33.99 & 35.93 & 36.50   & 28.11 & 36.12 & 34.39 & 36.68  & 26.48 \\
			0.4    & 37.54       & \textbf{39.16}      & 30.51 & 31.98 & 31.75   & 22.28 & 29.99 & 30.60 & 34.71  & 21.09 \\
			0.5    & 32.38       & \textbf{34.02}       & 26.51 & 26.79 & 24.67   & 17.59 & 22.17 & 24.64 & 31.62  & 16.91 \\ \hline
			\multicolumn{11}{c}{MAE}                                                                              \\ \hline
			0.1    & \textbf{0.17}        & \textbf{0.17}        & 1.16  & 0.21  & 0.18    & 0.23  & 0.22  & 0.36  & 0.51   & 0.27  \\
			0.2    & 0.36        & \textbf{0.35}        & 1.75  & 0.47  & 0.40    & 0.62  & 0.49  & 0.72  & 0.79   & 0.77  \\
			0.3    & 0.59        & \textbf{0.57}        & 2.53  & 0.92  & 0.72    & 1.63  & 0.87  & 1.27  & 1.13   & 2.19  \\
			0.4    & 0.89        & \textbf{0.83}        & 3.82  & 1.66  & 1.27    & 4.30  & 1.62  & 2.14  & 1.52   & 5.86  \\
			0.5    & 1.52        & \textbf{1.34}        & 6.25  & 3.37  & 3.36    & 10.54 & 4.45  & 4.41  & 2.13   & 13.58 \\ \hline
			\multicolumn{11}{c}{SSIM}                                                                             \\ \hline
			0.1    & \textbf{0.997}       & \textbf{0.997}       & 0.952 & 0.992 & 0.995   & 0.993 & 0.993 & 0.991 & 0.983  & 0.987 \\
			0.2    & 0.993       & \textbf{0.994}       & 0.905 & 0.979 & 0.989   & 0.960 & 0.986 & 0.977 & 0.977  & 0.938 \\
			0.3    & 0.988       & \textbf{0.990}       & 0.850 & 0.946 & 0.973   & 0.835 & 0.973 & 0.948 & 0.968  & 0.774 \\
			0.4    & 0.977       & \textbf{0.984}      & 0.762 & 0.878 & 0.934   & 0.583 & 0.925 & 0.884 & 0.958  & 0.498 \\
			0.5    & 0.932       &\textbf{0.954}       & 0.615 & 0.733 & 0.773   & 0.317 & 0.694 & 0.714 & 0.941  & 0.275 \\ \toprule[2pt]
			\multicolumn{11}{c}{CRAYONS}                                                                          \\ \toprule[2pt]
			\multicolumn{11}{c}{PSNR {[}dB{]}}                                                                    \\ \hline
			0.1    & 41.32       & \textbf{41.42}       & 40.55 & 38.65 & 39.69   & 37.83 & 36.11 & 36.11 & 37.10  & 37.71 \\
			0.2    & 37.90       & \textbf{38.13}       & 37.07 & 35.25 & 36.23   & 33.23 & 33.37 & 33.37 & 35.49  & 32.77 \\
			0.3    & 35.45       & \textbf{35.99}       & 34.11 & 32.81 & 33.58   & 28.45 & 31.27 & 31.27 & 33.69  & 27.90 \\
			0.4    & 33.16       & \textbf{34.02}      & 31.06 & 30.12 & 30.00   & 23.63 & 28.69 & 28.69 & 31.98  & 23.22 \\
			0.5    & 30.03       & \textbf{30.97}       & 27.48 & 26.54 & 25.33   & 19.54 & 25.58 & 25.58 & 29.62  & 19.32 \\ \hline
			\multicolumn{11}{c}{MAE}                                                                              \\ \hline
			0.1    & \textbf{0.40}        & \textbf{0.40}        & 1.10  & 0.51  & 0.46    & 0.52  & 0.79  & 0.79  & 1.23   & 0.54  \\
			0.2    & 0.83        & \textbf{0.82}       & 1.65  & 1.06  & 0.93    & 1.17  & 1.45  & 1.45  & 1.65   & 1.27  \\
			0.3    & 1.30        & \textbf{1.27}        & 2.39  & 1.74  & 1.50    & 2.26  & 2.25  & 2.25  & 2.16   & 2.54  \\
			0.4    & 1.87        & \textbf{1.79}       & 3.53  & 2.71  & 2.37    & 4.52  & 3.39  & 3.39  & 2.74   & 5.15  \\
			0.5    & 2.75        & \textbf{2.58}        & 5.60  & 4.46  & 4.28    & 9.01  & 5.34  & 5.34  & 3.58   & 10.17 \\ \hline
			\multicolumn{11}{c}{SSIM}                                                                             \\ \hline
			0.1    & \textbf{0.991}       & \textbf{0.991}      & 0.981 & 0.983 & 0.987   & 0.982 & 0.979 & 0.979 & 0.964  & 0.978 \\
			0.2    & 0.980       & \textbf{0.981}       & 0.961 & 0.962 & 0.972   & 0.946 & 0.955 & 0.955 & 0.951  & 0.936 \\
			0.3    & 0.967       & \textbf{0.969}       & 0.928 & 0.931 & 0.951   & 0.851 & 0.920 & 0.920 & 0.934  & 0.832 \\
			0.4    & 0.948       & \textbf{0.954}      & 0.865 & 0.875 & 0.906   & 0.666 & 0.858 & 0.858 & 0.914  & 0.645 \\
			0.5    & 0.904       & \textbf{0.918}       & 0.742 & 0.760 & 0.785   & 0.436 & 0.742 & 0.742 & 0.884  & 0.433 \\ \hline
		\end{tabular}
	\end{center}
\end{table*}

\begin{table*}[]\small
	\caption{Comparison of the denoising efficiency of the proposed network for impulsive noise removal with the state-of-the-art methods on the test dataset \cite{Mal1}. \label{tab:test_final}} 
	\begin{center}
		
		\begin{tabular}{lllllllllll}  \hline \hline
			$\rho$ & IDCNN\textsubscript{BSD500} & IDCNN\textsubscript{GoogleV4} & DnCNN          & FAPGF & FASTAMF & FFNRF & FRF   & INRF & PARIGI & PGF          \\ \hline \hline
			\multicolumn{11}{c}{Average PSNR {[}dB{]}}                                                                    \\ \hline
			0.1         & 40.12       & \textbf{40.45} & 38.92 & 36.78   & 37.97 & 36.20 & 37.22 & 34.82  & 34.33 & 36.68 \\
			0.2         & 37.02       & \textbf{37.36} & 36.18 & 34.11   & 35.06 & 32.38 & 33.88 & 32.60  & 32.75 & 31.88 \\
			0.3         & 34.77       & \textbf{35.26} & 33.64 & 31.80   & 32.53 & 27.77 & 31.08 & 30.62  & 31.15 & 27.20 \\
			0.4         & 32.68       & \textbf{33.38} & 30.86 & 29.22   & 29.33 & 23.20 & 27.28 & 28.16  & 29.52 & 22.84 \\
			0.5         & 29.92       & \textbf{30.51} & 27.52 & 25.92   & 24.77 & 19.20 & 21.71 & 24.73  & 27.78 & 19.09 \\ \hline
			\multicolumn{11}{c}{Average MAE}                                                                              \\ \hline
			0.1         & 0.46        & \textbf{0.45}  & 1.18  & 0.72    & 0.60  & 0.72  & 0.59  & 1.01   & 1.62  & 0.63  \\
			0.2         & 0.92        & \textbf{0.90}  & 1.71  & 1.31    & 1.12  & 1.34  & 1.21  & 1.65   & 2.18  & 1.40  \\
			0.3         & 1.42        & \textbf{1.39}  & 2.41  & 2.02    & 1.72  & 2.44  & 2.01  & 2.45   & 2.82  & 2.75  \\
			0.4         & 2.01        & \textbf{1.94}  & 3.49  & 3.04    & 2.60  & 4.78  & 3.30  & 3.59   & 3.58  & 5.50  \\
			0.5         & 2.86        & \textbf{2.75}  & 5.45  & 4.80    & 4.59  & 9.48  & 6.56  & 5.66   & 4.53  & 10.75 \\ \hline
			\multicolumn{11}{c}{Average SSIM}                                                                             \\ \hline
			0.1         & 0.989       & \textbf{0.989} & 0.980 & 0.977   & 0.983 & 0.975 & 0.981 & 0.974  & 0.949 & 0.976 \\
			0.2         & 0.978       & \textbf{0.979} & 0.962 & 0.955   & 0.967 & 0.940 & 0.961 & 0.949  & 0.931 & 0.929 \\
			0.3         & 0.963       & \textbf{0.966} & 0.932 & 0.920   & 0.944 & 0.841 & 0.935 & 0.913  & 0.908 & 0.815 \\
			0.4         & 0.944       & \textbf{0.949} & 0.872 & 0.856   & 0.898 & 0.645 & 0.875 & 0.848  & 0.881 & 0.624 \\
			0.5         & 0.904       & \textbf{0.911} & 0.748 & 0.736   & 0.771 & 0.414 & 0.666 & 0.718  & 0.845 & 0.420 \\ \hline
		\end{tabular}
		
	\end{center}
\end{table*}

\section{CONCLUSIONS}

In this work, we introduced a switching filter that employs deep neural networks for impulsive noise removal in color images.  The performed experiments revealed that the proposed network architecture which operates on a modified version of DnCNN network for impulse detection and adaptive mean filter for impulses restoration allows to efficiently remove impulses and outperforms state-of-the-art filters in terms of PSNR, MAE and SSIM measures.  Future work will be focused on modification of the proposed neural network architecture to detect and restore pixels contaminated by impulsive noise in a single stage network instead of using impulses detection and restoration as separate denoising operations.

\section*{Acknowledgment} 
This work was supported by the Polish National Science Centre under the project 
2017/25/B/ST6/02219,
and was also funded by the Statutory Research funds of  Silesian University of Technology, Poland (Grant BK/200/Rau1/2019).

\ifCLASSOPTIONcaptionsoff
  \newpage
\fi



%
\bibliographystyle{IEEEtran}
\bibliography{IEEEabrv,fast-bibliografia-revised} 

%




\end{document}